\documentclass[lettersize,journal]{IEEEtran}
\usepackage{amsmath,amsfonts}
\usepackage{amssymb}
\usepackage{algorithmic}
\usepackage{algorithm}
\usepackage{array}
\usepackage[caption=false,font=normalsize,labelfont=sf,textfont=sf]{subfig}
\usepackage{textcomp}
\usepackage{stfloats}
\usepackage{float}
\usepackage{url}
\usepackage{verbatim}
\usepackage{graphicx}
\usepackage{cite}
\usepackage{epsfig}
\usepackage{acronym}
\usepackage{bm}
\usepackage{epstopdf}
\usepackage{multirow}
\usepackage{hyperref}
\usepackage{xcolor}
\usepackage{booktabs}
\usepackage{multicol}
\usepackage{booktabs}
\usepackage{color}
\usepackage{soul}

\usepackage{float}
\newcommand{\etal}{\textit{et al}. }

\begin{document}


\title{AG-ReID.v2: Bridging Aerial and Ground Views for Person Re-identification}

\author{\IEEEauthorblockN{Huy Nguyen, Kien Nguyen,  Sridha Sridharan, Clinton Fookes}\\
\IEEEauthorblockA{\textit{School of Electrical Engineering and Robotics} \\
\textit{Queensland University of Technology, Brisbane, Australia 4000} \\
thanhnhathuy.nguyen@hdr.qut.edu.au, \{k.nguyenthanh, s.sridharan, c.fookes\}@qut.edu.au}}

\markboth{IEEE Transactions on Information Forensics and Security, 2023}
{Nguyen \MakeLowercase{\textit{\etal}}: AG-ReID.v2: Bridging Aerial and Ground Views for Person Re-identification}


\maketitle

\begin{abstract}
Aerial-ground person re-identification (Re-ID) presents unique challenges in computer vision, stemming from the distinct differences in viewpoints, poses, and resolutions between high-altitude aerial and ground-based cameras. Existing research predominantly focuses on ground-to-ground matching, with aerial matching less explored due to a dearth of comprehensive datasets. To address this, we introduce AG-ReID.v2, a dataset specifically designed for person Re-ID in mixed aerial and ground scenarios. This dataset comprises 100,502 images of 1,615 unique individuals, each annotated with matching IDs and 15 soft attribute labels. Data were collected from diverse perspectives using a UAV, stationary CCTV, and smart glasses-integrated camera, providing a rich variety of intra-identity variations. Additionally, we have developed an explainable attention network tailored for this dataset. This network features a three-stream architecture that efficiently processes pairwise image distances, emphasizes key top-down features, and adapts to variations in appearance due to altitude differences. Comparative evaluations demonstrate the superiority of our approach over existing baselines. We plan to release the dataset and algorithm source code publicly, aiming to advance research in this specialized field of computer vision. For access, please visit \href{https://github.com/huynguyen792/AG-ReID.v2}{https://github.com/huynguyen792/AG-ReID.v2}.
\end{abstract}


\section{Introduction}

Person Re-identification (ReID) is a technique in computer vision that identifies and matches individuals across images or videos captured by multiple, non-overlapping cameras \cite{Ye2021DeepLF,BeyondIntra,Zheng2016PersonRP,Liu2020PairbasedUA,Li2019DomainAwareUC}. This method is advantageous in surveillance systems as it does not depend on high-resolution biometric data, such as facial features, which are often required for more precise identification methods. Person ReID has diverse applications including video surveillance, retail management, search and rescue operations, healthcare services, and public safety initiatives. These applications demonstrate the utility of person ReID in enhancing safety measures and optimizing resource distribution in various environments.

\begin{figure}
    \centering
    
    \begin{tabular}{c}
        \includegraphics[width=0.5\columnwidth, height=3cm]{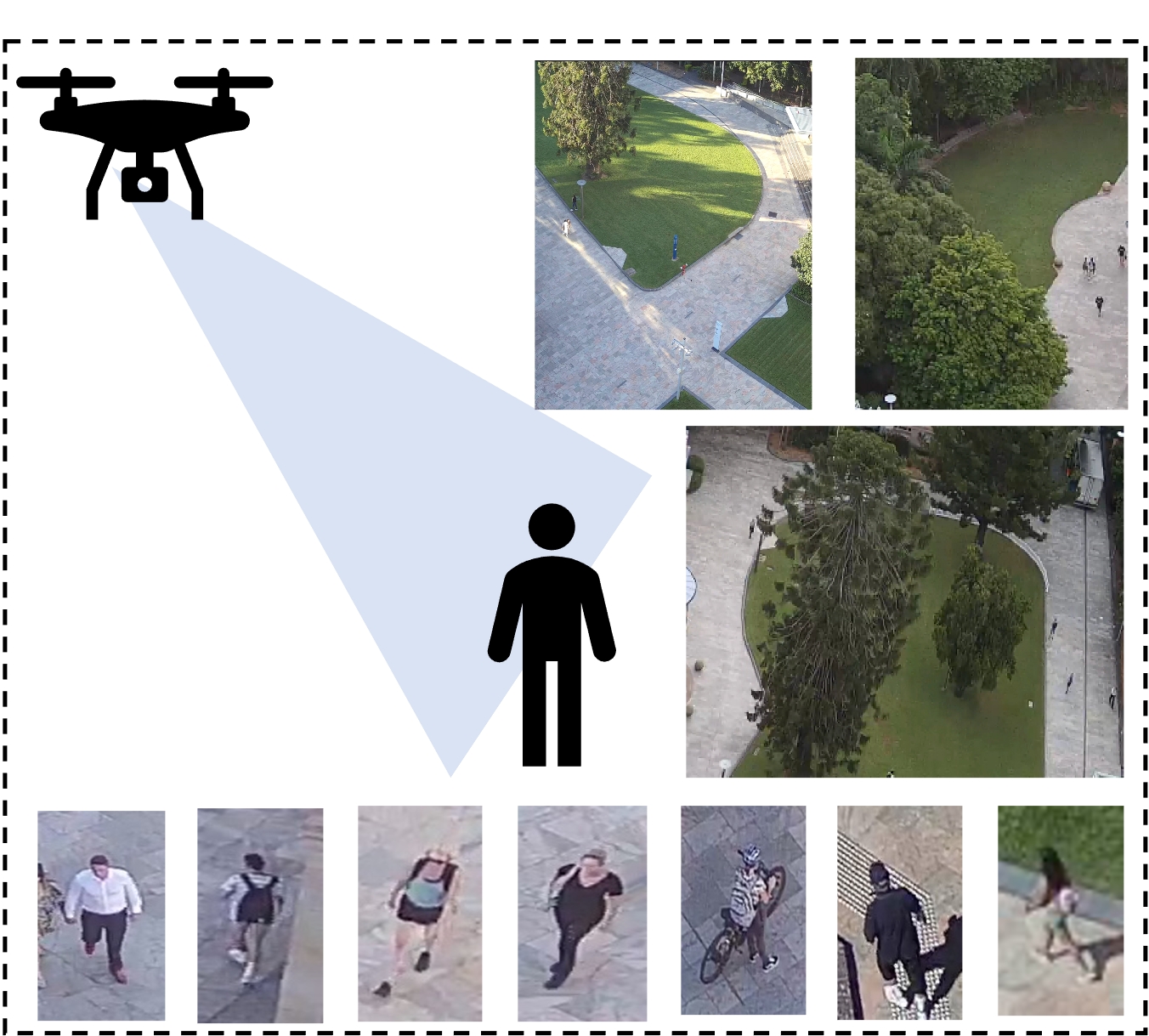} \\
        \small (a) Aerial camera.                                                 
    \end{tabular}
    
    \begin{multicols}{2} 
        \begin{tabular}{@{}c@{}}
            \includegraphics[width=1\columnwidth, height=3cm]{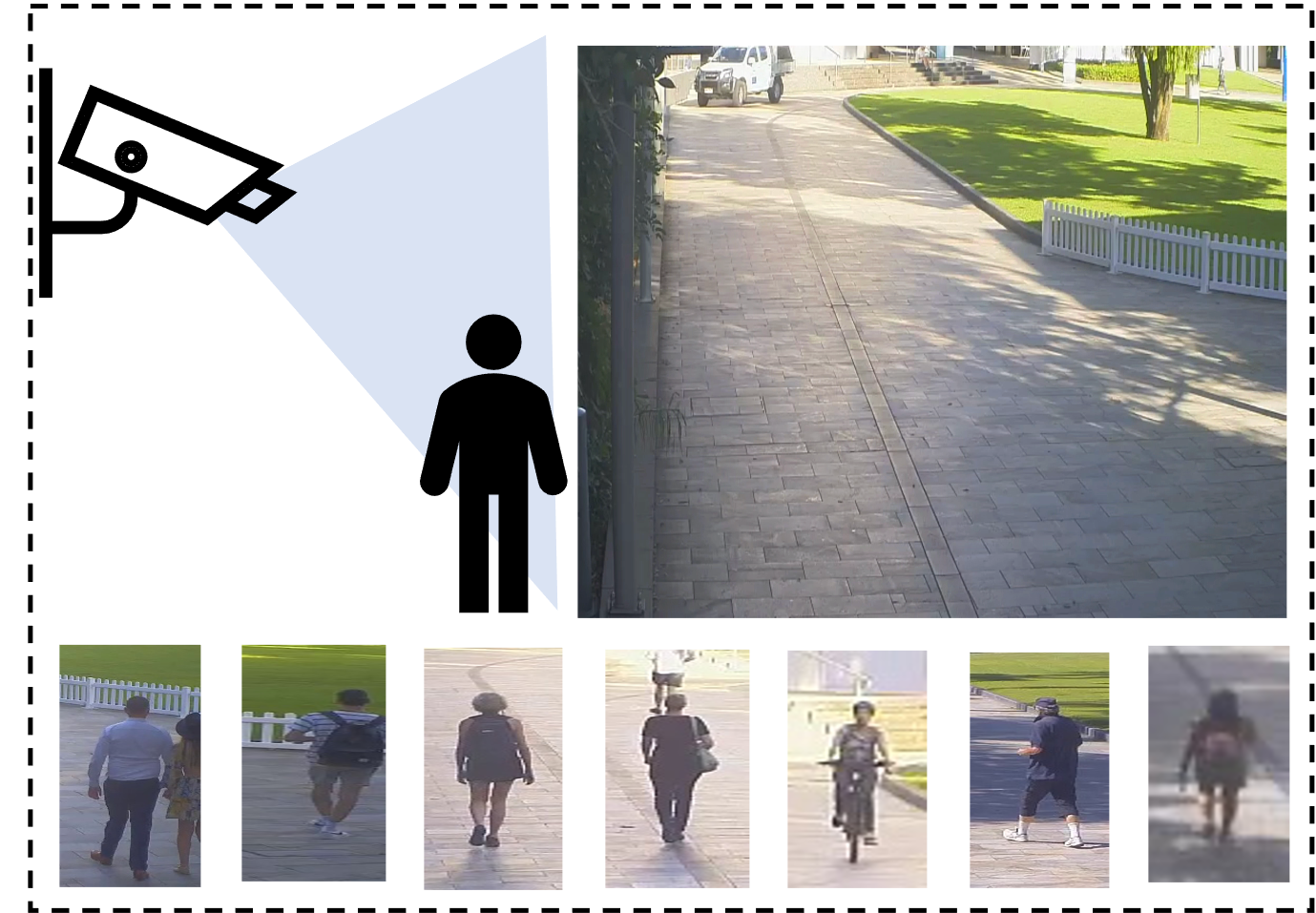} \\
            \small (b) CCTV camera.                                              
        \end{tabular}
        
        \begin{tabular}{@{}c@{}}
            \includegraphics[width=1\columnwidth, height=3cm]{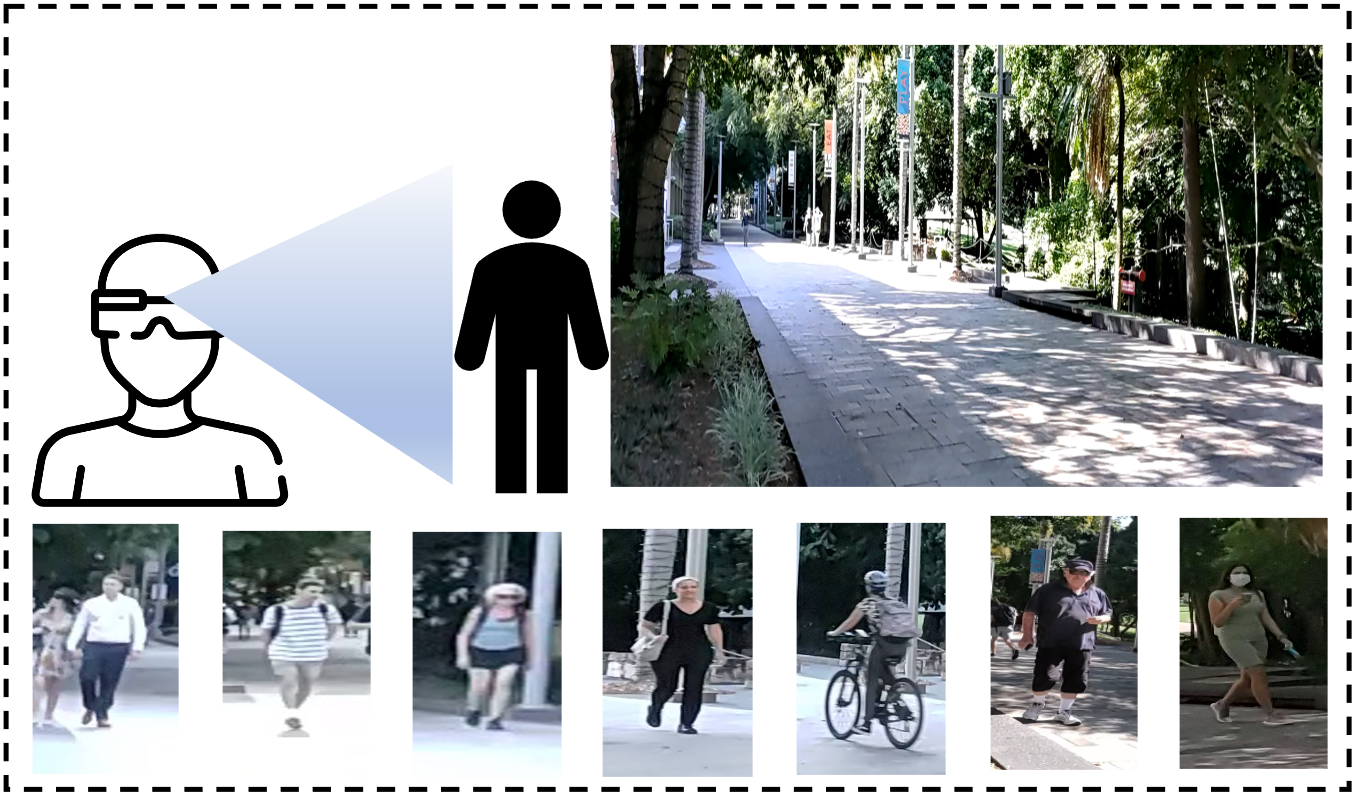} \\
            \small (c) Wearable camera.                                                
        \end{tabular}
    \end{multicols}
    
    \caption{Aerial (a), CCTV (b), and wearable camera (c) perspectives vary in resolution, occlusion, and lighting in the AG-ReID.v2 dataset.}
    \label{fig:perspective}
\end{figure}

The development of advanced airborne platforms and imaging sensors has significantly advanced the field of aerial person ReID \cite{AerialSurveillance}. These platforms, in comparison with traditional ground-based systems, offer notable advantages, such as increased scale and mobility, and the ability to perform both covert and overt observations \cite{Li2021UAVHumanAL, Yu2020ScaleMF}. High-altitude aerial cameras are particularly effective, capable of capturing extensive areas with reduced occlusion and demonstrating versatility in various operational conditions \cite{Granelli2020ADA, Singh2018EyeIT, Soldi2021SpaceBasedGM}. The integration of multi-modal sensors, including visual, thermal, and LiDAR technologies, further enhances the effectiveness of aerial ReID, improving target visibility and robustness \cite{greenwood_2021, Soldi2021SpaceBasedGM}.

Traditionally, research in aerial person ReID has primarily focused on matching within aerial imagery \cite{Kumar2021ThePA, Li2021UAVHumanAL, Zhang2019PersonRI}. However, the integration of aerial and ground images for person ReID poses distinct challenges, largely due to differences in viewpoints, poses, and resolutions. This area, while crucial, lacks comprehensive and publicly available datasets suitable for such cross-domain studies. Existing datasets, such as the one by Schumann \etal \cite{Schumann2017PersonRA}, face limitations in terms of accessibility, scale, and diversity of collection environments.

Addressing these challenges, this paper introduces the AG-ReID.v2 dataset, an extension of our previous AG-ReID.v1 dataset \cite{nguyen2023aerialground}. AG-ReID.v2 encompasses a broader range of aerial and ground imagery, providing a diverse, publicly accessible resource for ReID research. This dataset, meticulously compiled, reflects a variety of real-world scenarios, particularly suited for surveillance and monitoring applications.

Furthermore, we propose a novel three-stream architecture for aerial-ground person ReID, designed to address the specific challenges in this domain. Each stream within our model focuses on particular aspects such as localized attributes and soft-biometric markers, enabling a more precise and interpretable matching process. This innovative approach represents a significant step forward in addressing the complexities of aerial-ground person ReID. Images from the AG-ReID.v2 dataset, illustrating these aspects, are shown in Figure \ref{fig:perspective}.

The dataset comprises various image types and modalities, including:
\begin{itemize}
    \item Aerial images were obtained using DJI M600 Pro drones, equipped with XT2 sensors and $24$mm lenses., flown at altitudes from $15$ to $45$ meters. These images offer an elevated perspective of scenes and individuals, aiding in person identification and tracking from an aerial viewpoint.
    \item Ground-based CCTV images taken by a standard low-resolution CCTV camera positioned at key locations on a campus. These provide a traditional ground-level perspective, similar to standard CCTV systems.
    \item Images from a wearable camera, attached to smart glasses, providing a first-person viewpoint. These high-resolution images capture dynamic details like head and body posture, as well as contextual information about the environment and surrounding elements.
\end{itemize}

This diverse collection of aerial and ground-based images forms a comprehensive dataset suitable for developing and testing person ReID models in aerial-ground settings. The aerial images, taken from higher altitudes, introduce unique challenges in ReID due to differences in viewpoint, pose, and resolution, which affect the appearance of individuals compared to ground-level views. Ground images are sourced from two types of cameras—CCTV and wearable—offering variability in resolution, perspective, mobility, and lighting conditions. This diversity accurately reflects the complexities encountered in real-world surveillance and person ReID scenarios.

The AG-ReID.v2 dataset incorporates 15 soft-biometric attributes per individual, as illustrated in Figure \ref{fig:attribute}. These attributes cover various aspects including age, gender, and clothing style. They serve as supplementary information to facilitate attribute recognition and improve Re-identification (ReID) system efficacy.

In this work, we introduce a novel method for person re-identification (ReID) to accompany our dataset, employing a three-stream architecture tailored for the specific challenges posed by integrating aerial and ground images. This architecture emphasizes local features, such as the head region and soft-biometric attributes, to enhance the ReID process's explainability. The model incorporates a simplified localization layer in the elevated-view attention stream, opting out of using complex pose estimators. This layer functions as an adaptive mechanism, dynamically shifting focus between global and head-specific features in response to the input. Such an approach enables the system to discern and prioritize essential features for accurate person re-identification. This method not only bolsters performance in scenarios involving aerial and ground ReID but also provides insights into the pivotal attributes that drive the identification process.

In summary, the key contributions of our study are:

\begin{itemize}
\item \textit{Introduction of the AG-ReID.v2 dataset:} The AG-ReID.v2 dataset integrates images from both aerial (UAV) and ground sources (CCTV and wearable cameras), providing a more comprehensive view compared to datasets focused solely on ground or aerial perspectives. This dataset includes a large volume of images representing a significant number of unique identities and is enhanced with 15 soft attributes, contributing to its depth and applicability for diverse analytical purposes.

\item \textit{A three-stream person ReID model with an explainable elevated-view attention mechanism:} We present a novel three-stream architecture tailored for the challenges specific to aerial-ground person ReID. This model features an elevated-view attention mechanism to address aerial-ground perspective challenges, and an explanation component for visualizing appearance differences, thereby augmenting the model's interpretability.

\item \textit{Comprehensive experimental analysis:} An extensive evaluation of our model using the AG-ReID.v2 dataset demonstrates its effectiveness, showing improvement over existing ReID models. This analysis underscores the practical utility of our approach in aerial-ground person ReID scenarios.

\item \textit{Public dataset and code release:} In an effort to support research in this field, we are making the AG-ReID.v2 dataset and the code for our baseline person ReID system available to the public.
\end{itemize}

The structure of this paper is organized into several sections: Section \ref{Related Work} provides a review of the relevant literature. Section \ref{AG-ReID.v2 Dataset} details the AG-ReID.v2 dataset. In Section \ref{Three-Stream Aerial-Ground ReID}, we introduce our proposed method, focusing on an explainable approach for aerial-ground re-identification. Section \ref{Experiments} reports on the experimental setup and results. Finally, Section \ref{Conclusion} summarizes our findings and contributions.

\begin{figure}
    \centering
    \includegraphics[width=0.95\columnwidth]{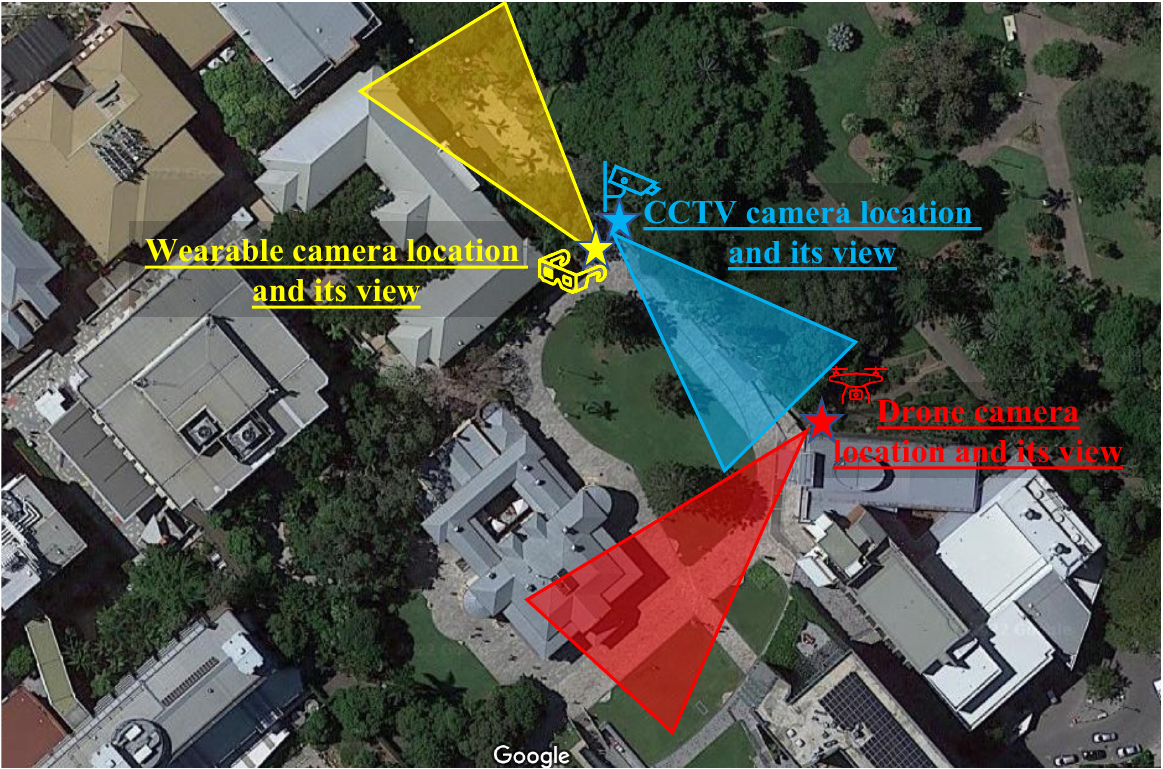}
    \caption{Data Collection Areas for the AG-ReID.v2 dataset. }
    \label{fig:DataCollectionArea}
\end{figure}

\section{Related Work}
\label{Related Work}

\subsection{Datasets for Person ReID}

In the field of Person Re-identification (ReID), a variety of datasets have been developed, each with unique characteristics suited for different research scenarios:

\begin{itemize}
    \item \textbf{Ground-Ground Person ReID Datasets}: Ground-ground datasets are commonly used in ReID studies. For example, Market-1501 \cite{Zheng2015ScalablePR}, established in 2015, contains 32,668 images from 1,501 individuals. In comparison, DukeMTMC-reID \cite{Gou2017DukeMTMC4ReIDAL} (retracted in June 2019) comprises 36,411 images but represents a slightly smaller pool of 1,404 individuals. These datasets illustrate the varying scopes and scales prevalent in ground-based ReID research.
    
    \item \textbf{Aerial-Aerial Person ReID Datasets}: The interest in aerial-based ReID datasets is growing. PRAI-1581 \cite{Zhang2019PersonRI}, released in 2019, includes 39,461 images of 1,581 subjects. Another significant contribution is the UAV-Human dataset \cite{Li2021UAVHumanAL}, introduced in 2021. It encompasses 41,290 images of 1,144 individuals, captured by a drone flying at altitudes between 2 to 8 meters over various locations and times. This dataset's versatility makes it a valuable resource for a range of surveillance applications.

    \item \textbf{Aerial-Ground Person ReID Dataset}: To address the evolving needs of surveillance technologies, datasets combining aerial and ground views have been developed. Our earlier work, AG-ReID.v1 \cite{nguyen2023aerialground}, focused on integrating UAV and CCTV camera perspectives. Extending this, the current study introduces AG-ReID.v2, an enhanced dataset that surpasses existing datasets in terms of diversity and scale. AG-ReID.v2 features 100,502 images of 1,615 unique individuals, captured using CCTV, UAVs, and wearable devices. The images, recorded from heights of 15 to 45 meters, include 15 distinct attributes per individual. Table \ref{tab:compare_statics} provides a comprehensive comparison of AG-ReID.v2 with other datasets, highlighting its unique contributions to the ReID research landscape.
\end{itemize}

\begin{table*}
    \centering
    \fontsize{8}{10}\selectfont
    \caption{Comparison of AG-ReID.v2 with other ReID datasets. Our AG-ReID.v2 leads in identities, images and platforms.}
    \begin{tabular}{lcccccc}
        \toprule
        \textbf{Attributes} & \multicolumn{2}{c}{\textbf{Ground-Ground}} & \multicolumn{2}{c}{\textbf{Aerial-Aerial}} & \multicolumn{2}{c}{\textbf{Aerial-Ground}} \\ 
        \cmidrule(r){2-3} \cmidrule(lr){4-5} \cmidrule(l){6-7}
        & Market-1501 & DukeMTMC-reID & PRAI-1581 & UAV-Human & AG-ReID.v1 & AG-ReID.v2 (ours) \\  
        \midrule
        \# IDs & 1,501 & 1,404 & 1,581 & 1,144 & 388 & \textbf{1,615}\\ 
        \# Images & 32,668 & 36,411 & 39,461 & 41,290 & 21,983 & \textbf{100,502}\\ 
        \# Attributes & $\times$ & $\times$ & $\times$ & 7 & 15 & 15\\ 
        Backgrounds & $\times$ & $\times$ & $\times$ & $\checkmark$ & $\checkmark$ & $\checkmark$\\ 
        Occlusion & $\times$ & $\times$ & $\checkmark$ & $\checkmark$ & $\checkmark$ & $\checkmark$\\ 
        Camera Views & fixed & fixed & mobile & mobile & mixed & mixed \\ 
        Platforms & CCTV & CCTV & UAV & UAV & Dual & \textbf{Triple} \\ 
        Altitude & $<10m$ & $<10m$ & $20\sim60m$ & $2\sim8m$ & $15\sim45m$ & $15\sim45m$\\ 
        \# UAVs & 0 & 0 & 2 & 1 & 1 & 1\\ 
        \bottomrule
    \end{tabular}
    \label{tab:compare_statics}
\end{table*}

\subsection{Person ReID Approaches}

\textbf{Person Re-identification} in computer vision aims to consistently identify individuals across varied camera views. Key models in this domain include BoT \cite{Luo2019BagOT}, which integrates label smoothing, random erasing, and auto-augmentation, MGN \cite{Wang2018LearningDF}, characterized by its unique triplet loss function and multi-granularity feature utilization, and SBS \cite{Qi2021StrongerBF}, enhancing triplet loss for improved accuracy. These models typically employ backbone architectures such as ResNet \cite{He2016DeepRL}, OSNet \cite{yolov5-strongsort-osnet-2022}, and ViT \cite{Dosovitskiy2021AnII}, and are evaluated on datasets like Market-1501, DukeMTMC-reID, and MSMT17. The advent of aerial-aerial person ReID, supported by datasets such as PRAI-1581 \cite{Zhang2019PersonRI} and UAV-Human \cite{Li2021UAVHumanAL}, has introduced new approaches, including subspace pooling \cite{Zhang2019PersonRI} and the DG-NET framework \cite{Zheng2019JointDA} \cite{Li2021UAVHumanAL}. Our research extends these foundations, combining HRNet-18 \cite{Wang2019DeepHR} and Swin Transformer models \cite{Liu2021SwinTH} \cite{liu2021swinv2}, aiming to enhance person re-identification capabilities.

\textbf{Aerial-Aerial Person ReID} The PRAI-1581 \cite{Zhang2019PersonRI} and UAV-Human \cite{Li2021UAVHumanAL} datasets have catalyzed new algorithmic developments for aerial-aerial matching. Zhang \etal \cite{Zhang2019PersonRI} introduced subspace pooling to generate concise, discriminative features for aerial ReID. Zheng \etal's DG-NET \cite{Zheng2019JointDA}, applied to the UAV-Human dataset \cite{Li2021UAVHumanAL}, focuses on enhancing ReID embeddings through a joint learning framework that combines ReID learning with synthetic data generation.

\textbf{Multi-stream Person ReID Architecture} Multi-stream architectures are increasingly utilized in person re-identification, addressing various aspects of the matching challenge. For instance, Chung \etal \cite{Chung2017ATS} proposed a two-stream architecture, segregating spatial and temporal information learning. Xie \etal \cite{Xie2021LowresolutionAT} introduced a three-stream architecture that merges features from RGB, low-resolution, and grayscale images to tackle image quality variation. Khatun \etal \cite{Khatun2020JointIF} developed a four-stream architecture, applying dual identification and verification losses on four input images to optimize intra-class and inter-class distances. These multi-stream designs illustrate the effectiveness of integrating different data types or features in enhancing the accuracy and robustness of person re-identification systems.

\section{AG-ReID.v2 Dataset}
\label{AG-ReID.v2 Dataset}

\subsection{{Dataset Collection}}

Our dataset is collected on a university campus using three cameras: a UAV, a CCTV, and a wearable camera. Each operates in distinct non-overlapping areas, as illustrated in Figure \ref{fig:DataCollectionArea}. Specifically, the UAV camera captures aerial images in the red-marked area, the CCTV camera records ground images in the blue area, and the wearable camera operates within the yellow area.

In the AG-ReID.v2 dataset, data collection was conducted using a variety of recording devices to capture a comprehensive range of real-world pedestrian activities. The DJI M600 Pro UAV, equipped with a DJI XT2 camera, was designated to record video in the red area, offering a high 4K resolution at 30 frames per second (FPS). In a different setting, the Bosch CCTV camera was responsible for capturing footage in the blue area, with a resolution of 800 x 600 pixels at the same frame rate of 30 FPS. {Additionally, the person wearing the Vuzix M4000 smart glasses, who was positioned in the yellow area, remained stationary during the recording sessions. This stationary stance was a strategic decision to ensure consistent and stable image quality, particularly critical for the pedestrian subjects being recorded. The stable recording minimizes the potential variables that could arise from the movement of the camera operator, thereby maintaining the clarity and consistency of the image quality}. Such high-quality recording is essential for the effectiveness of pedestrian detection and recognition models, which are a central objective of our dataset. By utilizing this diverse array of recording methods and equipment, the AG-ReID.v2 dataset provides a robust and realistic platform for training and enhancing machine learning models for person re-identification. The specific details of the camera equipment used, including brands, models, resolutions, frame rates, and recording zones, are detailed in Table \ref{camera_specs} for further reference.

\begin{table}
    \centering
    \fontsize{8}{10}\selectfont
    \caption{Specifications of Cameras Used for Data Collection}
    \begin{tabular}{lccccc}
        \toprule
        \textbf{Device} & \textbf{Brand} & \textbf{Model} & \textbf{Resolution} & \textbf{FPS} & \textbf{Altitude}   \\ 
        \midrule
        CCTV & Bosch & N/A & $800$ $\times$ $600$ & $30$ & $\approx 3m$ \\
        Wearable & Vuzix & M4000 & 4K & $30$ & $\approx 1.5m$  \\ 
        UAV & DJI & XT2 & $3840$ $\times$ $2160$ & $30$ & ~ $15\sim45m$   \\ 
        \bottomrule
    \end{tabular}
    \label{camera_specs}
\end{table}

{The AG-ReID.v2 dataset was methodically compiled over a period of 5 months, comprising  20 distinct data collection sessions. These sessions were not held consecutively; instead, they were strategically planned on non-consecutive days, allowing for adjustments based on varying weather conditions and area restrictions. To maintain uniform environmental conditions, the data collection was structured into specific time windows: Morning sessions were conducted from $8:45$ am to $10:00$ am, and afternoon sessions from $3:30$ pm to $4:45$ pm. This schedule was deliberately chosen to capture the unique atmospheric and lighting conditions prevalent in early mornings and late afternoons.}

During each session, UAV flights were conducted at three different altitudes—15m, 35m, and 45m Above Ground Level (AGL)—with each flight lasting 15 minutes. This systematic approach was instrumental in collecting a wide array of both aerial and ground-based images, capturing pedestrians in diverse lighting scenarios such as sunny versus rainy weather and direct sunlight versus shade. The dataset robustly addresses real-world challenges encountered in pedestrian detection and recognition, including issues related to occlusion, blur, resolution, and viewpoint variations. It also comprises images depicting pedestrians in various states of motion and captured from multiple angles, thereby significantly enhancing the dataset's utility for developing and refining pedestrian detection and recognition models. These models are specifically designed to navigate the complexities of real-world environments, addressing challenges like motion blur, occlusion, varying resolutions, and diverse viewpoints, as highlighted in the literature \cite{wang2022feature, AmenaPose}. The intermittent and meticulously timed nature of these data collection sessions was crucial in ensuring the dataset’s diversity, making it a valuable asset for advancing research in pedestrian detection and recognition technologies.

In total, the dataset includes 100,502 images and 1,615 unique identities. The use of multiple cameras and altitudes provides a wide range of variations in viewpoint, lighting, and background, making it an ideal dataset for evaluating the robustness and generalization of person re-identification models.

\begin{figure}
    \centering
    \includegraphics[width=0.9\linewidth,height=5cm,keepaspectratio]{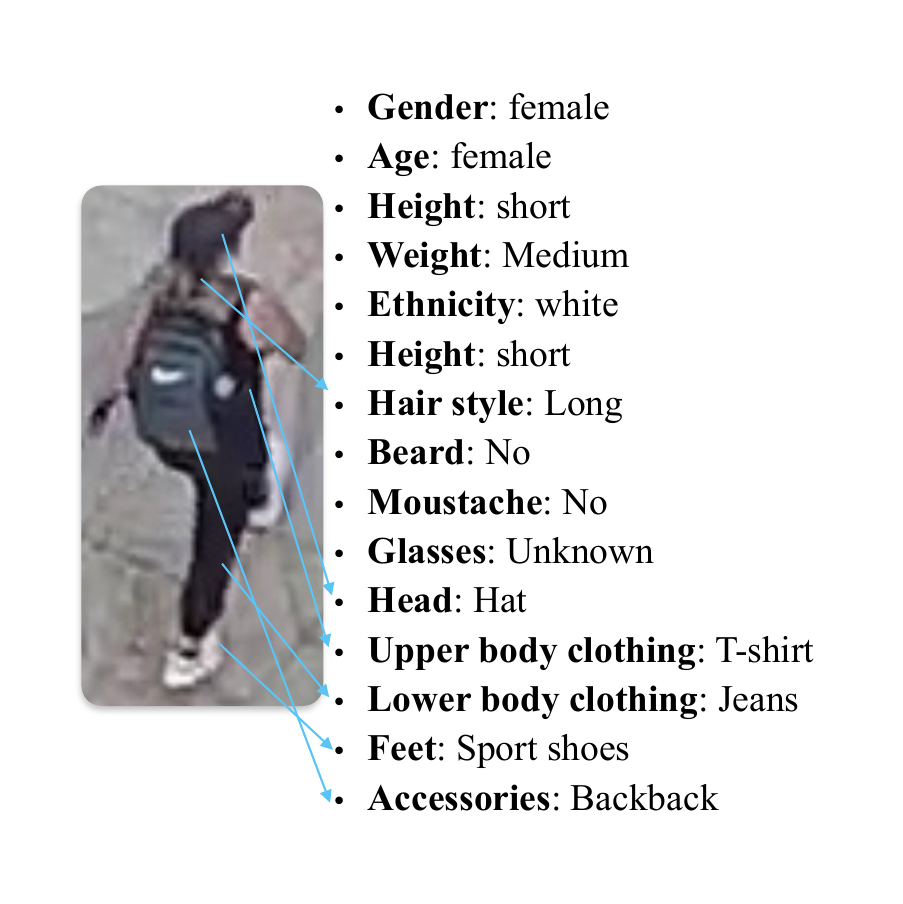}
    \vspace{-0.4cm}
    \caption{15 soft-biometric labels in the AG-ReID.v2 dataset.}
    \label{fig:attribute}
    \vspace{-0.2cm}
\end{figure}

\subsection{{Annotations}}

\begin{figure}
  \center
   \includegraphics[width=0.95\linewidth]{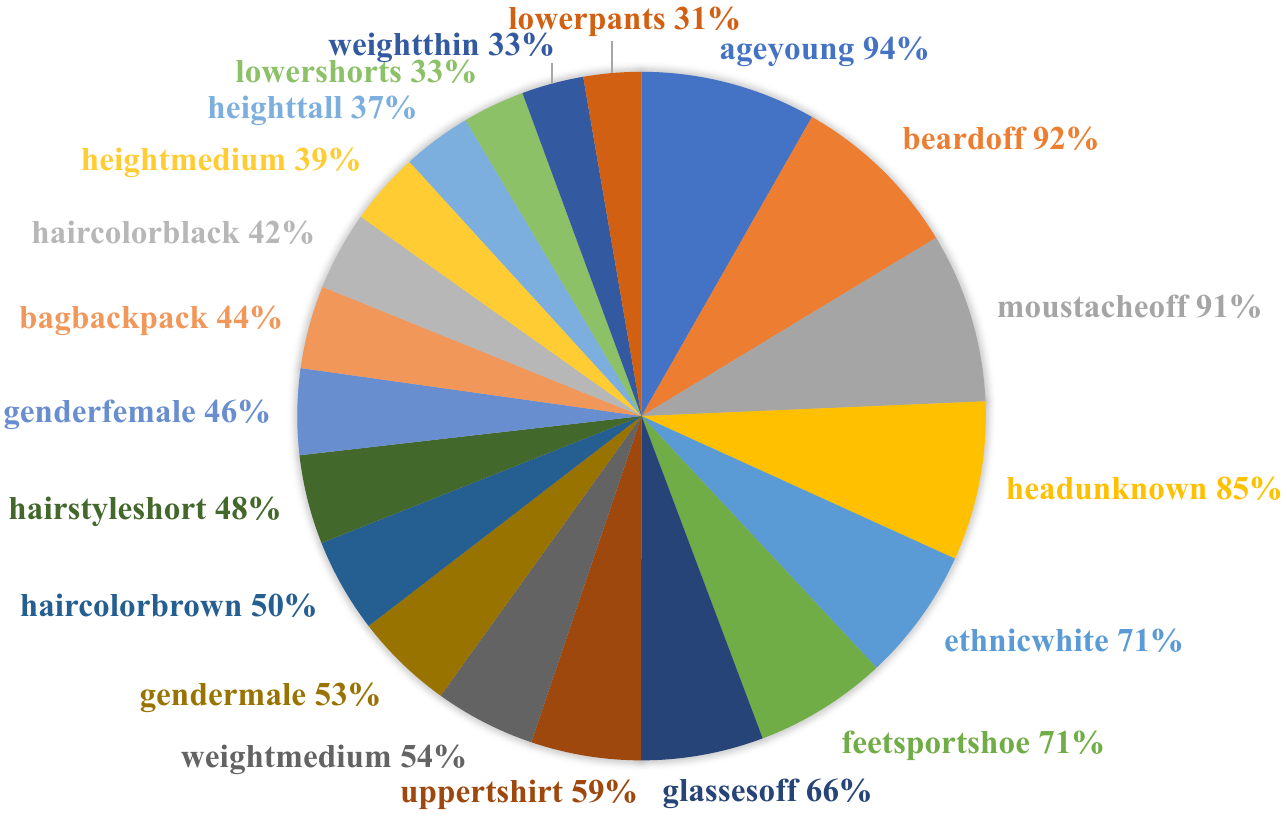}\\
  \caption{Top 20 attribute distribution in our dataset.}\label{attributes_top_20}
\end{figure}

\begin{figure}
\centering
  \includegraphics[width=1\linewidth]{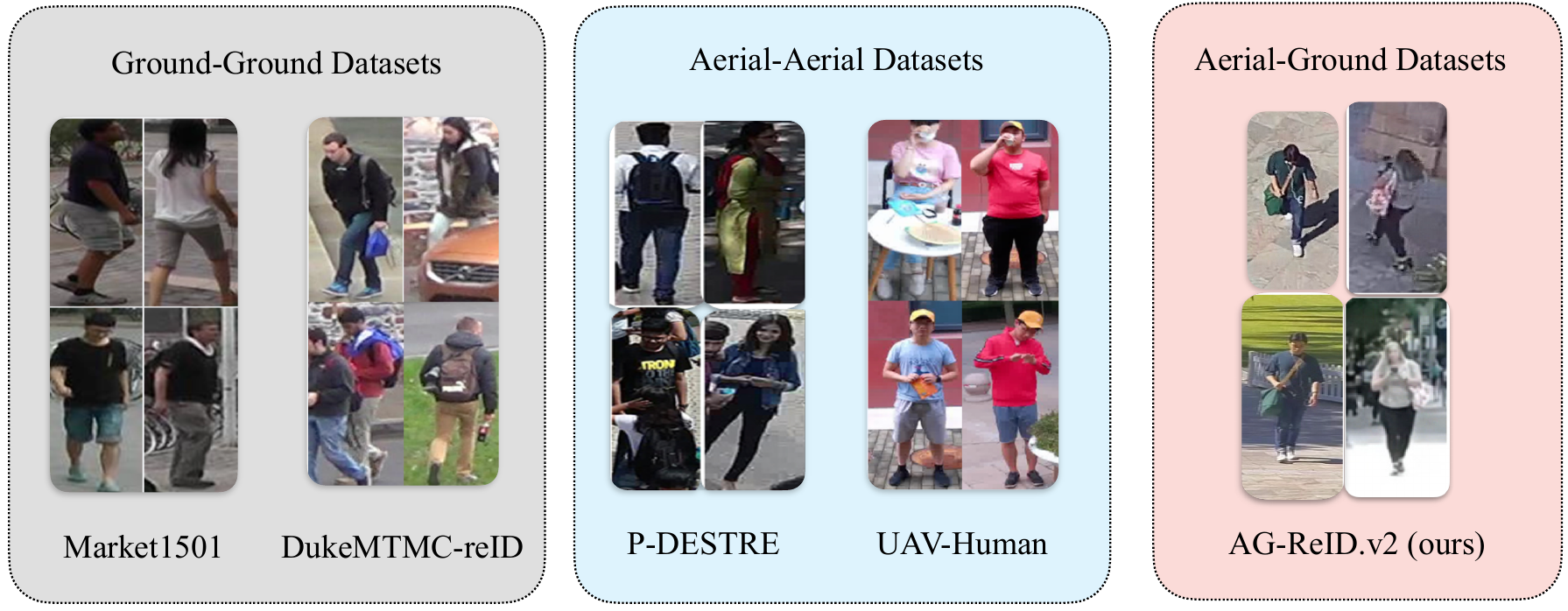} 
  \caption{{Example images} from two ground-ground datasets, Market-1501 \cite{Zheng2015ScalablePR} and DukeMTMC-reID \cite{Gou2017DukeMTMC4ReIDAL}, alongside two aerial-aerial datasets, P-DESTRE \cite{Kumar2021ThePA} and UAV-Human \cite{Li2021UAVHumanAL}, in comparison with our aerial-ground dataset, AG-ReID.v2. The images from AG-ReID.v2 highlight distinct challenges associated with reconciling perspective variances between ground-based (bottom row) and aerial-based (top row) images of individuals. This contrast is not as prevalent in the other datasets, which are confined to a single domain, either aerial or ground.}
  \label{fig:dataset_comp}
\end{figure}

In the AG-ReID.v2 dataset, we utilize the YOLO detector \cite{li2022yolov6} and the StrongSORT tracker \cite{yolov5-strongsort-osnet-2022} for person detection and tracking in video data, saving an image every 30 frames. {Manual correction is applied to address inaccuracies, such as incorrectly identified objects. The dataset's primary focus is on short-term re-identification from various platforms, including ground and aerial perspectives, without tracking individuals over extended periods. Thus, changes in clothing or accessories are not a primary consideration in this dataset. }Annotators match and label individuals across cameras, and the dataset includes manual attribute annotations, as indicated in Figure \ref{fig:attribute}. These attributes are based on 15 soft-biometric labels from \cite{Kumar2021ThePA, Lin2019ImprovingPR}, relevant due to their UAV data collection similarities. The dataset's top 20 attributes are presented in Figure \ref{attributes_top_20}, and a comparative analysis with other public datasets is shown in Table \ref{tab:compare_statics}, along with exemplar images in Figure \ref{fig:dataset_comp}. 

{The AG-ReID.v2 dataset's selection of 15 attributes is informed by a thorough comparative analysis with existing datasets in the ground, aerial, and aerial-ground categories. This analysis, detailed in Table} \ref{tab:attribute_comparison}, {highlights the unique aspects of AG-ReID.v2 and the rationale behind its attribute selection.}

\begin{table*}
    \centering
    \fontsize{7}{8}\selectfont
    \caption{{Comparative Analysis of Attributes in AG-ReID.v2 and Other Datasets}}
    \begin{tabular}{lcccccc}
        \toprule
        \textbf{Attributes} & \multicolumn{2}{c}{\textbf{Ground-Ground}} & \multicolumn{2}{c}{\textbf{Aerial-Aerial}} & \multicolumn{2}{c}{\textbf{Aerial-Ground}} \\ 
        \cmidrule(r){2-3} \cmidrule(lr){4-5} \cmidrule(l){6-7}
        & Market-1501 & DukeMTMC-reID & P-DESTRE & UAV-Human & AG-ReID.v1 & AG-ReID.v2 (ours) \\  
        \midrule
        Gender & \textcolor{green}{$\checkmark$} & \textcolor{green}{$\checkmark$} & \textcolor{green}{$\checkmark$} & \textcolor{green}{$\checkmark$} & \textcolor{green}{$\checkmark$} & \textcolor{green}{$\checkmark$}\\ 
        Age & \textcolor{red}{$\times$} & \textcolor{red}{$\times$} & \textcolor{green}{$\checkmark$} & \textcolor{red}{$\times$} & \textcolor{green}{$\checkmark$} & \textcolor{green}{$\checkmark$}\\ 
        Height & \textcolor{red}{$\times$} & \textcolor{red}{$\times$} & \textcolor{green}{$\checkmark$} & \textcolor{red}{$\times$} & \textcolor{green}{$\checkmark$} & \textcolor{green}{$\checkmark$}\\ 
        Body Volume & \textcolor{red}{$\times$} & \textcolor{red}{$\times$} & \textcolor{green}{$\checkmark$} & \textcolor{red}{$\times$} & \textcolor{green}{$\checkmark$} & \textcolor{green}{$\checkmark$}\\ 
        Ethnicity & \textcolor{red}{$\times$} & \textcolor{red}{$\times$} & \textcolor{green}{$\checkmark$} & \textcolor{red}{$\times$} & \textcolor{green}{$\checkmark$} & \textcolor{green}{$\checkmark$}\\ 
        Hair Color & \textcolor{red}{$\times$} & \textcolor{red}{$\times$} & \textcolor{green}{$\checkmark$} & \textcolor{red}{$\times$} & \textcolor{green}{$\checkmark$} & \textcolor{green}{$\checkmark$}\\ 
        Hairstyle & \textcolor{red}{$\times$} & \textcolor{red}{$\times$} & \textcolor{green}{$\checkmark$} & \textcolor{red}{$\times$} & \textcolor{green}{$\checkmark$} & \textcolor{green}{$\checkmark$}\\ 
        Beard & \textcolor{red}{$\times$} & \textcolor{red}{$\times$} & \textcolor{red}{$\times$} & \textcolor{red}{$\times$} & \textcolor{green}{$\checkmark$} & \textcolor{green}{$\checkmark$}\\ 
        Moustache & \textcolor{red}{$\times$} & \textcolor{red}{$\times$} & \textcolor{red}{$\times$} & \textcolor{red}{$\times$} & \textcolor{green}{$\checkmark$} & \textcolor{green}{$\checkmark$}\\ 
        Glasses & \textcolor{red}{$\times$} & \textcolor{red}{$\times$} & \textcolor{green}{$\checkmark$} & \textcolor{red}{$\times$} & \textcolor{green}{$\checkmark$} & \textcolor{green}{$\checkmark$}\\ 
        Head Accessories & \textcolor{red}{$\times$} & \textcolor{red}{$\times$} & \textcolor{green}{$\checkmark$} & \textcolor{red}{$\times$} & \textcolor{green}{$\checkmark$} & \textcolor{green}{$\checkmark$}\\ 
        Upper Body Clothing & \textcolor{green}{$\checkmark$} & \textcolor{green}{$\checkmark$} & \textcolor{red}{$\times$} & \textcolor{green}{$\checkmark$} & \textcolor{green}{$\checkmark$} & \textcolor{green}{$\checkmark$}\\ 
        Lower Body Clothing & \textcolor{green}{$\checkmark$} & \textcolor{green}{$\checkmark$} & \textcolor{red}{$\times$} & \textcolor{green}{$\checkmark$} & \textcolor{green}{$\checkmark$} & \textcolor{green}{$\checkmark$}\\ 
        Feet & \textcolor{red}{$\times$} & \textcolor{red}{$\times$} & \textcolor{green}{$\checkmark$} & \textcolor{red}{$\times$} & \textcolor{green}{$\checkmark$} & \textcolor{green}{$\checkmark$}\\ 
        Accessories & \textcolor{green}{$\checkmark$} & \textcolor{green}{$\checkmark$} & \textcolor{green}{$\checkmark$} & \textcolor{green}{$\checkmark$} & \textcolor{green}{$\checkmark$} & \textcolor{green}{$\checkmark$}\\ 
        \bottomrule
    \end{tabular}
    \label{tab:attribute_comparison}
\end{table*}

The AG-ReID.v2 dataset incorporates a diverse range of attributes, covering physical characteristics, appearance, and accessories, to ensure comprehensive coverage suitable for various applications. The dataset balances ground-level details with aerial-specific requirements, making it ideal for cross-domain applications. Unique attributes like beard and moustache offer finer granularity in person re-identification, especially in ground-level images. Inclusion of ethnicity and a broad range of age categories enhances the dataset's applicability across diverse demographics. The selection of attributes like footwear type and accessories aligns well with the requirements of aerial-ground integrated surveillance systems, combining detailed close-up information with features discernible from a distance.

\subsection{Dataset Features}

\textbf{Diverse Identities}. The dataset is characterized by a comprehensive range of identities, represented in images captured using three types of cameras: an aerial camera and two ground-based cameras, specifically CCTV and wearable cameras. This combination results in significant differences between aerial and ground views, mirroring real-world scenarios. The dataset's unique configuration presents a complex challenge in cross-matching individuals, a task that is more straightforward in conventionally captured datasets.

\begin{figure}
  \center
  \includegraphics[width=0.5\linewidth]{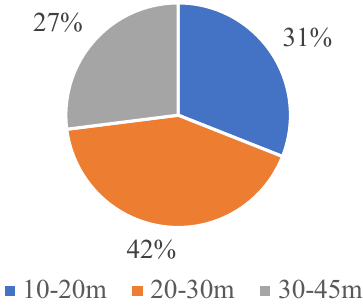}\\
  \caption{The distribution of imagery data across UAV flying altitudes.}\label{flying_altitude}
\end{figure}

\textbf{Variations in Altitude}. The AG-ReID.v2 dataset incorporates images captured at various altitudes, ranging from 15 to 45 meters, using Unmanned Aerial Vehicles (UAVs). This range in altitude leads to a broad spectrum of image scales and offers diverse perspectives of subjects from elevated positions. The impact of this altitude diversity on the imagery is illustrated in Figure \ref{flying_altitude}.

\begin{figure}[ht]
	\centering
	\includegraphics[width=0.95\linewidth]{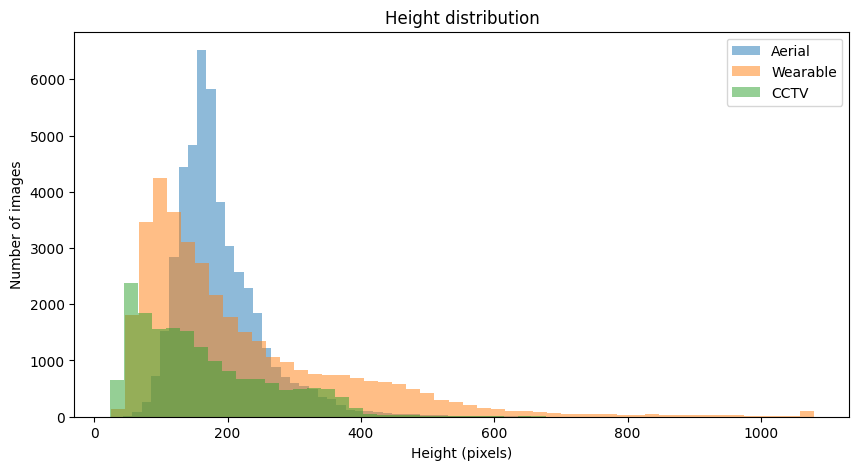}
	\caption{Distributions of the body heights (in pixels) across three cameras (aerial, wearable, CCTV) in the AG-ReID.v2 dataset.}
	\label{fig:height_width_distribution}
\end{figure}

\textbf{Resolution diversity}. The variations in camera resolutions and distances between the camera and subjects lead to distinct size differences in the cropped images of individuals. For instance, UAV-captured images range between 31x59 and 371x678 pixels, while CCTV-derived crops span from 22x23 to 172x413 pixels. Notably, the wearable camera produces images with dimensions comparable to both the UAV and CCTV. The dataset mostly features smaller-sized crops, making person reID particularly challenging. This variability is illustrated in Figure \ref{fig:height_width_distribution}, displaying body sizes recorded by the three cameras. The UAV captures sizes from 43 to 739 pixels, wearable cameras document from 25 to 1080 pixels, and CCTV images range from 23 to 622 pixels.

\begin{figure}
					
	\centering
	\begin{multicols}{2}				  
	\begin{tabular}{@{}c@{}}
		\includegraphics[width=1\columnwidth, height=1cm]{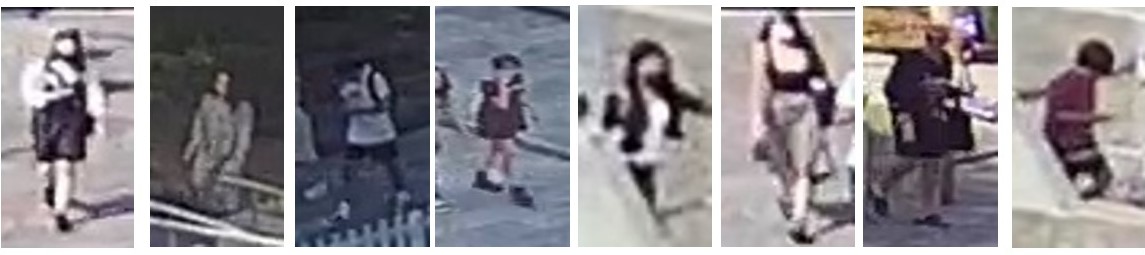} \\
		\small (a) Low Resolution                                                  
	\end{tabular}
					
	\begin{tabular}{@{}c@{}}
		\includegraphics[width=1\columnwidth, height=1cm]{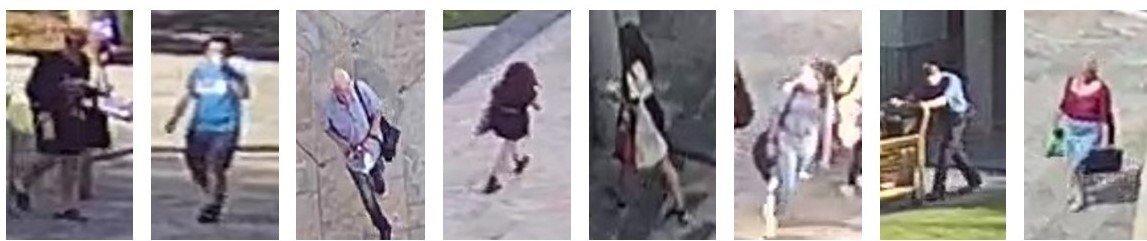} \\
		\small (b) Blur                                                  
	\end{tabular}
           
  \end{multicols}	
            		 \begin{multicols}{2} 
	\begin{tabular}{@{}c@{}}
		\includegraphics[width=1\columnwidth, height=1cm]{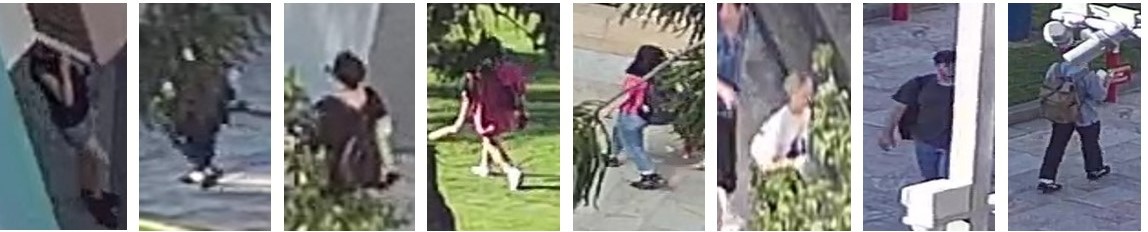} \\
		\small (c) Partially Occluded                                                  
	\end{tabular}
					
	\begin{tabular}{@{}c@{}}
		\includegraphics[width=1\columnwidth, height=1cm]{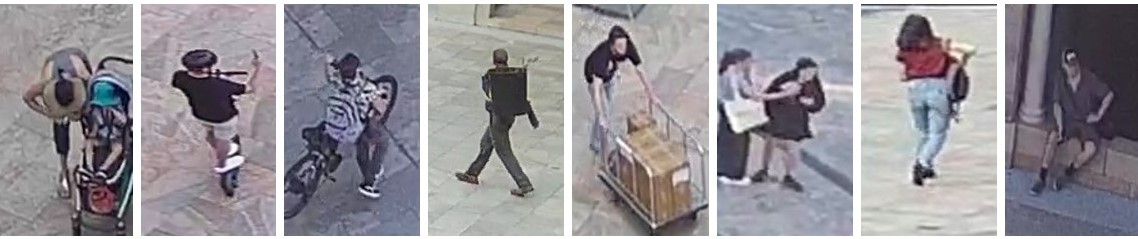} \\
		\small (d) Pose                                                  
	\end{tabular}
					   \end{multicols}	
        	 \begin{multicols}{2} 
	\begin{tabular}{@{}c@{}}
		\includegraphics[width=1\columnwidth, height=1cm]{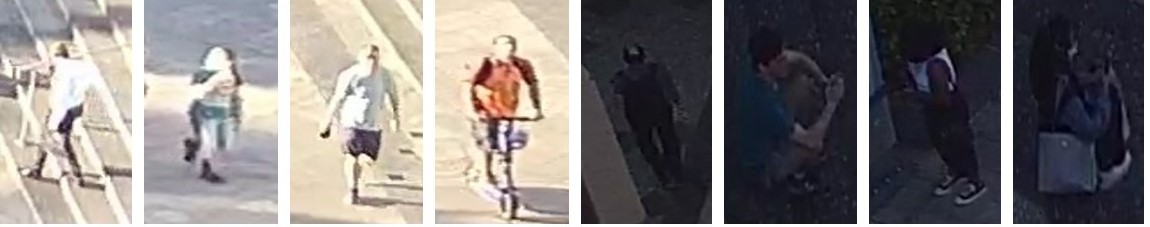} \\
		\small (e) Illumination                                                  
	\end{tabular}
					  
	\begin{tabular}{@{}c@{}}
		\includegraphics[width=1\columnwidth, height=1cm]{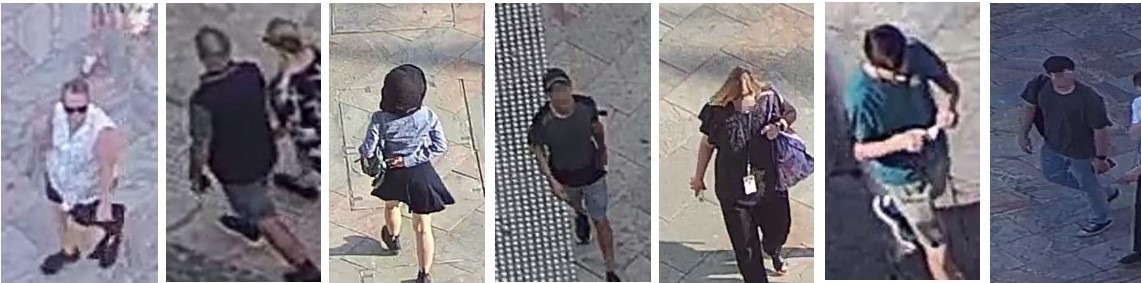} \\
		\small (f) Elevated Viewpoint                                             
	\end{tabular}
					  \end{multicols}
	\caption{Examples of the key challenges in the AG-ReID.v2 dataset.}
	\label{fig:Challenges}
\end{figure}

\textbf{Challenges}. Illustrated in Figure \ref{fig:Challenges}, our dataset introduces key challenges to person re-identification using UAVs. These include diverse scale variations due to the elevated UAV view, resolution variability where even 4K images may display low resolution at higher altitudes, and significant occlusions caused by trees and poles, or the presence of other individuals. Variable lighting conditions are another challenge, resulting from recordings at different times of the day, especially morning and afternoon. Additionally, the dataset captures noticeable motion blur from rapid subject movements and showcases an array of subject poses, from walking to biking, indicating the diversity of real-world scenarios. Our classification approach is visual-based: an image is categorized as “low resolution” if key features are indistinct, “occlusion” if a subject is significantly obscured, and “motion blur” if there is prominent blur from movement. This methodology aims to accurately represent the complexities of real-world UAV surveillance.

\subsection{Ethical Approval}
Our research team has secured ethics committee approval 
related to the "Multi-modal surveillance and video analytics" project, under the Human category. This authorization, valid until February 13, 2025, allows us to acquire and process video content with human subjects from both stationary and mobile cameras. To prioritize privacy, we have implemented facial pixelation on all captured videos, ensuring individual identities remain protected.

\section{{Three-Stream Aerial-Ground ReID}}\label{Three-Stream Aerial-Ground ReID}

\begin{figure*}
	\centering
	\includegraphics[width=0.95\linewidth]{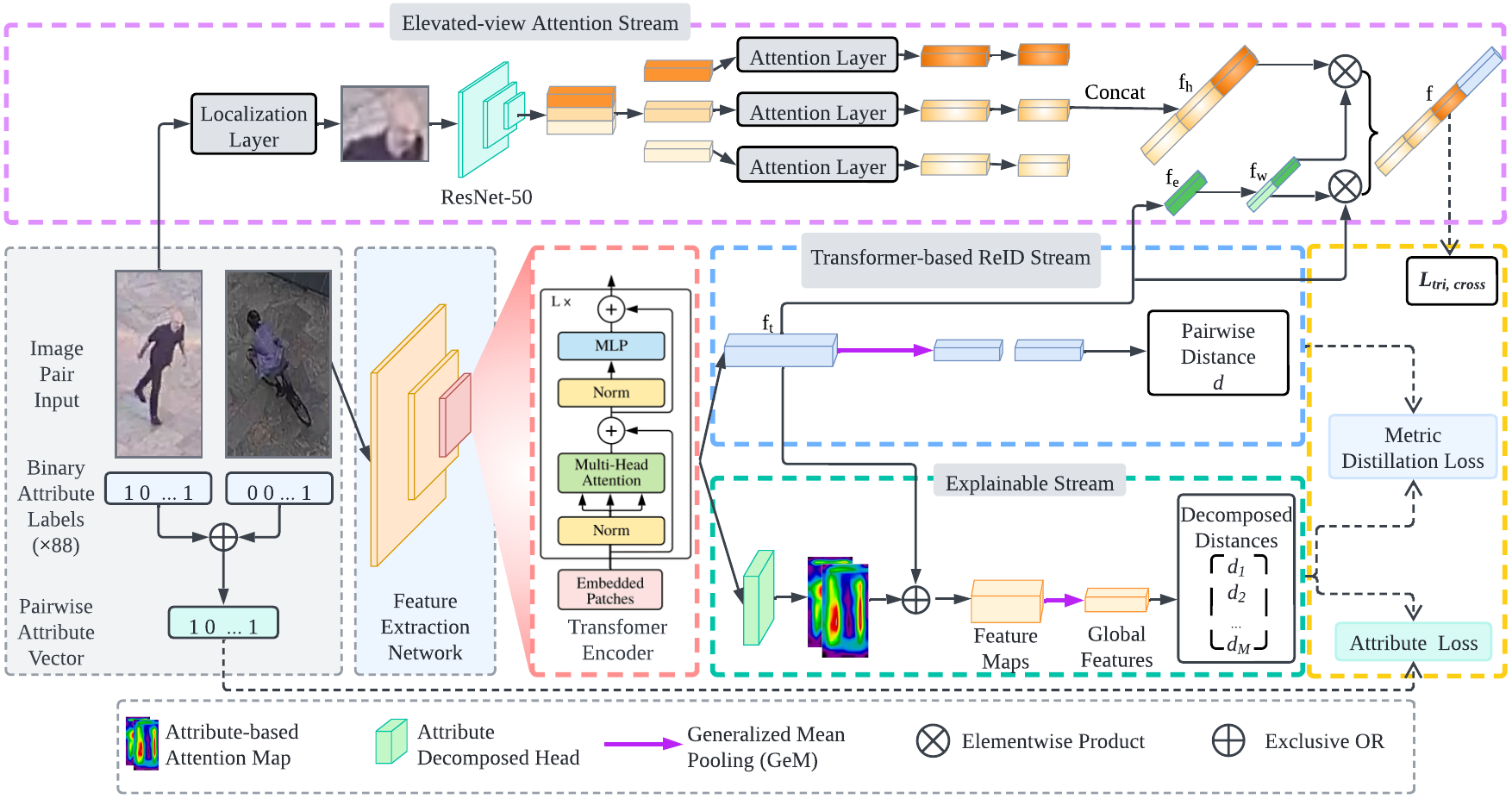}
	\caption{The architecture of our explainable elevated-view person ReID model utilizing a Vision Transformer (ViT) backbone is shown. The architecture comprises three streams: Stream 1 calculates the pairwise distance between input pair images, Stream 2 uses localization layers to crop the head region and extract local representations through attention layers, Stream 3 utilizes person attributes to account for variations in appearance due to varying flying altitudes and view angles between aerial and ground cameras. The final representation for ReID is generated by assembling the features from Stream 1 and the head features from Stream 2 through an adaptive module. Stream 3 is equipped with an Attribute Decomposed Head (ADH) module that produces attribute-guided attention maps for each attribute. }
	\label{fig:explainable_overal_architecture}
\end{figure*}

To address the complexities inherent in aerial-ground person ReID, our proposed model incorporates a three-stream architecture, as illustrated in Figure \ref{fig:explainable_overal_architecture}. This architecture consists of a transformer-based ReID stream for feature extraction, an elevated-view attention stream for detailed head region analysis, and an explainable ReID stream that utilizes attribute attention maps for refined feature representation. Stream 1 efficiently processes feature maps for discriminative analysis. Stream 2 focuses on augmenting head region features, crucial for analysis from aerial perspectives. In Stream 3, attribute feature maps are generated by amalgamating the output of Stream 1 with attribute attention maps, enhancing the model's interpretability. The model computes two distinct types of distances: metric distances derived from the features of Stream 1 and attribute-guided distances from Stream 3, both employing Generalized Mean Pooling (GeM). The model's training leverages cross-entropy and triplet losses, implemented in an integrated end-to-end approach. Further details of each stream are succinctly summarized in Table \ref{tab:three_stream_summary} and elaborated in subsequent sections.

\subsection{{Stream 1: Transformer-based Person ReID}}\label{subsec:Stream 1: Transformer-based Person ReID}
Our approach in developing a person ReID network, denoted as \(\mathcal{F(\cdot)}\), primarily utilizes the Stronger Baseline (SBS) model \cite{He2020FastReIDAP} as the foundational framework. The flexibility of our methodology allows for the integration of alternative models like MGN \cite{Wang2018LearningDF} or BOT \cite{Luo2019BagOT}, adapting to various application needs.

We chose the Vision Transformer (ViT) as our model's backbone due to its effectiveness in person ReID tasks. Introduced by Dosovitskiy \etal \cite{Dosovitskiy2021AnII}, ViT processes image patches as sequences, a method particularly apt for handling lower-resolution images common in aerial data. 

Other architectures like ResNet50 \cite{He2016DeepRL} and OSNet \cite{Zhou2019OmniScaleFL} were considered. ResNet50 is known for its deep residual learning, and OSNet for its efficient feature learning. However, ViT's ability to extract meaningful information from lower-resolution images made it the most suitable choice for our aerial imagery dataset.

Acknowledging existing ViT-based ReID methods like TransReid \cite{He2021TransReIDTO} and FED \cite{wang2022feature}, we noted their respective strengths and limitations. TransReid excels in global context but may lack in local feature extraction, while FED is effective against occlusions but less versatile in diverse datasets. Our model aims to balance global context and local feature extraction, offering robust performance across various scenarios.

The model's mechanics involve extracting feature maps \((F_i, F_j)\) from each image pair \((x_i, y_j)\). Using Generalized Mean Pooling (GMP), we convert these maps into feature vectors \((f_i, f_j)\), which form the basis for calculating pairwise distances \(d_{i,j}\). This metric is crucial in Stream 3 of our explainable ReID network for computing metric distillation loss, ensuring consistency across different streams of the network.

\subsection{{Stream 2: Elevated-view Attention Stream}}

{Our proposed method is designed to} enhance person re-identification techniques, specifically tailored for the AG-ReID.v2 dataset. {Building upon the work of }Xu \etal, in \cite{Xu2020BlackRA}. our model introduces a novel three-stream approach with a focus on an explainable model that accentuates the elevated-view perspective in person re-identification. This structured framework aims to provide clear and interpretable insights for improved identification from aerial viewpoints.

In our model, we adopt a different architecture from that of Xu \etal, which uses a ResNet50 backbone for a global stream. Instead, {our model utilizes the Vision Transformer Network}, which is specifically designed to incorporate an elevated-view stream. This stream includes a head feature extraction mechanism, supported by a localization layer inspired by the Spatial Transformer Network \cite{Jaderberg2015SpatialTN}. This localization layer is engineered to focus on the head region by performing spatial manipulations such as zooming, shifting, and cropping. These spatial transformations are defined by specific mathematical formulas that adjust the image to focus on the region of interest, as shown in the following equation:

\begin{equation}
    \begin{pmatrix}
    x_{i}^{s}\\ 
    y_{i}^{s}
    \end{pmatrix}
    =\begin{bmatrix}
    s_x & 0 & t_x \\ 
    0 & s_y & t_y
    \end{bmatrix}
    \begin{pmatrix}
    x_{i}^{t}\\ 
    y_{i}^{t}
    \end{pmatrix},
\end{equation}

where the location of each pixel $i^{th}$ within an image is determined by its original coordinates \((x_{i}^{s}, y_{i}^{s})\) and transformed coordinates \((x_{i}^{t}, y_{i}^{t})\). Through the use of scaling parameters \((s_x, s_y)\) and translation parameters \((t_x, t_y)\), we calibrate the image's scale and position. The region of interest, particularly the head, is normalized into a consistent shape to extract a feature map with dimensions \(C \times H \times W\). 

Our approach modifies traditional ReID methods by segmenting the feature map into three distinct horizontal sections. Each section is processed by a dedicated attention layer. The mathematical formulation of the attention layer is defined as follows:

\begin{equation}
    d_{i} = \sigma (U_{i}\text{ReLU}(W_{i}X_{i})),
\end{equation}

where \(\sigma\) represents the sigmoid activation function. In this context, \(U_{i}\) is responsible for increasing dimensionality, ReLU (Rectified Linear Unit) introduces non-linearity, \(W_{i}\) serves to reduce dimensions, and \(X_{i}\) denotes the respective input slice of the feature map.

The output of the attention layer, when applied to the input \(X_{i}\), is described by the equation:

\begin{equation}
    A_{i} =  X_{i} + X_{i} \cdot d_{i},
\end{equation}

which illustrates the element-wise multiplication of \(X_{i}\) with its attention scores \(d_{i}\). This computation produces an enhanced feature representation \(A_{i}\), combining the original input with its attention-modified counterpart.

The combined output from these layers, denoted as \(A_{i}\), leads to the formation of the final head feature \(f_{h}\), articulated as:

\begin{equation}
    f_{hi} = A_{i} \cdot  \xi \left(
\sum_{n=1}^{C} A_{i}^{n}
\right),
\end{equation}

where each channel \(n\) of the feature \(A_{i}\) is aggregated and subsequently multiplied by the channel-specific features to generate the head feature representation \(f_{h}\). This feature is further processed through a fully connected layer, resulting in an embedded feature \(f_{e}\) with dimensions \(N \times 2\), where \(N\) represents the batch size.

To assemble the final feature vector \(f\), we integrate the weighted head feature \(f_{h}\) with the base feature \(f_{t}\) from Stream 1 as follows:

\begin{equation}
\begin{cases}
        f = (f_{t} \cdot w_{1} ) \oplus (f_{h} \cdot w_{2} ) \\
        f_{w} = [w_{1} \hspace{1em} w_{2}]
\end{cases},
\end{equation}

with \(\oplus\) indicating the concatenation operation. The resultant feature vector \(f\) encompasses a comprehensive and refined representation of individuals, optimizing the effectiveness of the re-identification process.

\begin{table*}
    \centering
    \fontsize{8}{10}\selectfont
    \caption{{Streamlined Summary of Three-Stream Architecture in AG-ReID.v2}}
    \label{tab:three_stream_summary}
    \begin{tabular}{lccc}
        \toprule
        \textbf{Stream} & \textbf{Description} & \textbf{Function} & \textbf{Contribution} \\ 
        \midrule
        \textbf{Stream 1} & ViT for aerial image features. & Feature map extraction. & Metric consistency and distance. \\ 
        \midrule
        \textbf{Stream 2} & Head region focus with spatial manipulation. & Head feature enhancement. & Aerial view identification. \\ 
        \midrule
        \textbf{Stream 3} & ADH for feature refinement. & Attention maps for attributes. & Model interpretability and analysis. \\ 
        \bottomrule
    \end{tabular}
\end{table*}

\subsection{{Stream 3: Explainable ReID Stream}}

Our explainable ReID network, represented as \(\mathcal{G(\cdot)}\), is structurally similar to the conventional ReID model \(\mathcal{F(\cdot)}\), with both models utilizing initial layers that focus on key visual features such as texture and color, which are vital for recognizing attributes. The distinctiveness of \(\mathcal{G(\cdot)}\) is highlighted by integrating the Attribute Decomposition Head (ADH), which is an advancement over the traditional model architecture.

{This Attributes Decompose Head (ADH) is positioned after the final layer of the network $\mathcal{G(\cdot)}$, serving as a crucial component for feature refinement}. It begins with a convolutional layer of dimensions $\frac{C}{8}\times3\times3$, where $C$ is the number of channels from the preceding convolutional layer in $\mathcal{G(\cdot)}$. This is followed by a convolution of $M\times1\times1$, where $M$ denotes the number of attributes. The activation function $\delta(\cdot)$ is employed here, which is vital for the generation of attribute-guided attention maps (AAMs). These AAMs are instrumental in providing a deeper understanding of the model's focus on specific attributes, thereby enhancing the explainability of the network's decision-making process.

The activation function utilized in our model is defined as follows:
\begin{equation}
\delta(x) = \begin{cases}
\mathcal{K}\cdot(x+1)^{\mathcal{T}}, & \text{for } x > 0 \\
\mathcal{K}\cdot e^{x}, & \text{for } x \leq 0
\end{cases},
\label{equ:equ6}
\end{equation}
\noindent{%
\begin{minipage}{\columnwidth} where \(\mathcal{K}\)  and $\mathcal{T}$ is a scaling factor between 0 and 1, and \(x\) is the input from the preceding convolutional layer in the Attributes Decompose Head (ADH) module. This function is designed to enhance the model's focus on relevant attribute regions while mitigating potential biases.
\end{minipage}
}
Attribute Attention Maps (AAMs) are derived from the ADH module. Unlike general spatial attention mechanisms that target larger spatial areas, AAMs provide a detailed attention distribution geared towards individual attributes. These maps are represented as \(A_i\) and \(A_j\), with dimensions ${\in}R^{M\times w \times h}$, where each element \(A_i^k\) or \(A_j^k\) indicates the level of attention allocated to the \(k^{th}\) attribute.

The attribute-specific feature maps, denoted as \(F_i^k\) and \(F_j^k\), are formulated as:
\begin{equation}
\begin{cases}
F_i^k = F_i \bigotimes A^k \\
F_j^k = F_j \bigotimes A^k
\end{cases},
\end{equation}
where \(A_i^k\), \(A_j^k\) are the attribute attention maps, and \(\bigotimes\) signifies the element-wise multiplication. Each image yields \(M\) attribute-guided feature maps, with the \(k^{th}\) attribute enhancing relevant pixels while diminishing others. These feature maps are further refined into attribute-guided feature vectors \(f_i^k\), \(f_j^k\) through Generalized Mean Pooling (GMP), facilitating the measurement of attribute-specific distances.

\subsection{Loss Function for the Explainable Three-Stream ReID Network}

Our network employs a composite loss function, \( L \), which integrates various loss components tailored to different aspects of the model. This integrated loss is expressed as:

\begin{equation}
	L = L_d + \alpha L_{p1} + \beta L_{p2} + \alpha L_{triplet} + \beta L_{ce},
	\label{eq:total_loss}
\end{equation}

In this formulation, \( L_d \) is the Metric Distillation Loss, critical for distance metric learning. \( L_{p1} \) and \( L_{p2} \) collectively constitute the Attribute Prior Loss, addressing attribute-specific features. \( L_{triplet} \) represents the Triplet Loss, focusing on the relative distances between different data points. \( L_{ce} \), the Cross-Entropy Loss, aids in classification tasks. The parameters \( \alpha \) and \( \beta \) are used to balance these different loss components, ensuring an optimal combination for effective training.

\subsubsection{ID Losses}

\paragraph{Triplet Loss} 
The Triplet Loss function is designed to ensure that, for any given anchor sample, the distance to a positive sample (similar to the anchor) is smaller than the distance to a negative sample (dissimilar to the anchor) by a predefined margin \( p \). It is defined as:
\begin{equation}
	\mathcal{L}_{triplet}(i,j,k) = \max(p + d_{ij} - d_{jk}, 0),
\end{equation}
where \( d_{ij} \) and \( d_{jk} \) measure the metric distances between the respective samples.

\paragraph{Cross-Entropy Loss} 
The Cross-Entropy Loss is a widely used loss function for classification tasks. It quantifies the discrepancy between the predicted probability distribution of the model and the true distribution, represented as:
\begin{equation}
	\mathcal{L}_{ce} = -\frac{1}{N}\sum_{i=1}^{N} p_i \log{q_i},
\end{equation}
where \( N \) is the number of training samples, \( q_i \) is the predicted probability of the model for the \( i^{th} \) sample, and \( p_i \) represents the corresponding one-hot encoded true label.

\subsubsection{{Metric Distillation Loss}}
{Influenced by the work of Chen} \etal \cite{Chen2021ExplainablePR}, function \(\mathcal{G}(\cdot)\) decomposes the distance \(d_{i,j}\), calculated by the target model \(\mathcal{F}(\cdot)\), into components influenced by individual attributes. This decomposition is mathematically formulated as:
\begin{equation}
    d_{i,j} \approx \hat{d}_{i,j} = \sum_{k=1}^{M} d_{i, j}^k,
\end{equation}
where \(M\) denotes the total number of attributes, and \(d_{i, j}^k\) represents the distance between \(x_i\) and \(x_j\) influenced by the \(k^{th}\) attribute. The reconstructed distance \(\hat{d}_{i,j}\) is an approximation of \(d_{i,j}\) as calculated by our model.

We define the metric distillation loss, diverging from standard distillation techniques typically used for classification, as:
\begin{equation}
	L_d = |d_{i,j} - \sum^M_{k=1} d_{i,j}^k|,
	\label{eq:metric_loss}
\end{equation}
aiming to ensure consistency between the overall distance metrics generated by the target model and the attribute-influenced distances produced by our explainable model \(\mathcal{G}(\cdot)\).

\subsubsection{{Attribute Prior Loss}}
{The Attribute Prior Loss, inspired by Chen} \etal \cite{Chen2021ExplainablePR}, is designed to emphasize the unique attributes of individuals. This loss function is particularly effective in cases where attributes are only weakly labeled. It focuses on distinct features that set individuals apart, such as unique accessories, rather than common characteristics like similar clothing. For a pair of input images with attributes \((x_i, y_i, a_i)\) and \((x_j, y_j, a_j)\), we calculate the pairwise attribute vector \({a}_{i,j}\) as follows:
\begin{equation} 
{a}_{i,j} = {a}_i \oplus {a}_j, 
\label{equ:equ5} 
\end{equation} 
where \(\oplus\) indicates the Exclusive OR operation. This vector \({a}_{i,j}\) helps in identifying both the shared and unique attributes between \(x_i\) and \(x_j\).

The Attribute Prior Loss incorporates constraints related to the influence of exclusive and shared attributes, expressed as:
\begin{equation}
\begin{cases}
\sum_{e=1}^{M_E} \frac{d_{i,j}^e}{\tilde{d_{i,j}}} \geq \left(\frac{M_E}{M}\right)^{\vartheta} \\
\sum_{c=1}^{N-M_E} \frac{d_{i,j}^c}{\tilde{d_{i,j}}} \leq 1 - \left(\frac{M_E}{M}\right)^{\vartheta}
\end{cases}
\label{equ:equ14}
\end{equation}
Here, \(\vartheta\) is a balancing parameter, \(d_{i,j}^e\) represents the distance for exclusive attributes, \(d_{i,j}^c\) for shared attributes, \(M_E\) is the number of exclusive attributes, and \(M\) is the total number of attributes.

The Attribute Prior Loss is divided into two components: \( L_{p1} \) and \( L_{p2} \). This division allows for a detailed approach to accounting for the impact of both exclusive and shared attributes in the ReID process.

\begin{table}
\fontsize{8}{10}\selectfont
    \centering
    \caption{Statistics of the testing set for the AG-ReID.v2 dataset.}
    \label{tab:static_test_set}
    \begin{tabular}{llrr}
        \toprule
         \textbf{Cam} & \textbf{Subset}  & \textbf{IDs} & \textbf{Images} \\ 
         \midrule
        Aerial & Query & 534 & 2,356\\
        CCTV & Gallery & 534 & 6,347\\ 
        \addlinespace
        Aerial & Query & 519 & 2,209\\
        Wearable & Gallery & 519 & 12,912\\ 
        \addlinespace
        CCTV & Query & 534 & 1,811\\
        Aerial & Gallery & 534 & 14,362\\ 
        \addlinespace
        Wearable & Query & 519 & 2,340\\
        Aerial & Gallery & 519 & 12,568\\ 
         \bottomrule
    \end{tabular}
\end{table}

\paragraph{First Part \( L_{p1} \)} The expression for \( L_{p1} \) is given by:
\begin{equation}
	\begin{split}
		L_{p1} = \max \left(0, \left(\frac{M_E}{M}\right)^v - \sum_{e=1}^{M_E} \frac{d^e_{i,j}}{\hat{d}_{i,j}}\right) \\
  + \max \left(0, \sum ^ {M - M_E} _ {c=1} \frac{d^c_{i,j}}{\hat{d}_{i,j}} - 1 + \left(\frac{M_E}{M}\right)^v\right)
	\end{split},
	\label{eq:att_loss_p1}
\end{equation}
where \( M_E \) represents the number of exclusive attributes and \( M \) is the total number of attributes. This part of the loss function focuses on the proportional influence of exclusive attributes, ensuring that they have a more significant impact on the overall distance compared to common attributes.

\paragraph{Second Part \( L_{p2} \)} The equation for \( L_{p2} \) is:
\begin{equation}
	\begin{split}
		L_{p2} = \sum ^{M_E} _{e=1} \max \left(0, e^{-\lambda} \frac{\left(\frac{M_E}{M}\right)^v}{M_E} - \frac{d_{i,j}^e}{\hat{d}_{i,j}}\right) \\ 
  + \sum ^{M - M_E} _{c=1} \max \left(0, \frac{d_{i,j}^c}{\hat{d}_{i,j}}- e^{\lambda} \frac{1-\left(\frac{M_E}{M}\right)^v}{M-M_E}\right)
	\end{split},
	\label{eq:att_loss_p2}
\end{equation}
with the value of \( \lambda \) determined as:
\begin{equation}
	\lambda = \frac{1}{2} \ln \frac{M - M_E\left(\frac{M_E}{M}\right)^v}{M_E(1-\left(\frac{M_E}{M}\right)^v)}.
\end{equation}
This component sets specific thresholds for the contributions of each type of attribute, emphasizing the lower bound for exclusive attributes and the upper limit for common ones.

By using these two loss functions, \( L_{p1} \) and \( L_{p2} \), our model finely tunes the balance between the collective and individual influences of attributes, enhancing both accuracy and interpretability in person ReID tasks.

\section{Experiments}\label{Experiments}

In section \ref{sec:Dataset Partition}, we outline the process of splitting our dataset for training and evaluation. We then present the implementation details and evaluation metrics in Section \ref{sec:Implementation and Evaluation Metrics}. In Section \ref{sec:Comparison with State-of-the-art Models}, we evaluate popular baseline methods and state-of-the-art ReID models with various modalities datasets. Finally, in Section \ref{sec:Explainable Elevated-view Attention method on AG-ReID.v2 Dataset}, we report the performance of our proposed explainable elevated-view attention ReID method on the AG-ReID.v2 dataset.

\subsection{Dataset Partition}\label{sec:Dataset Partition}

The AG-ReID.v2 dataset is evenly divided between training and testing sets, both following a 1:1 ratio. The training set has 807 unique identities with 51,530 images. In contrast, the testing set features the next 808 identities, totaling 48,972 images. For testing, the data is split into two main categories: aerial-ground and ground-aerial, highlighting the focus on aerial-ground matching. Within these categories, each identity can have between one to six images from one camera as queries. The gallery set uses images from the other camera. This division method is inspired by the Market-1501 dataset but with slight changes in the number of images per identity. Detailed information about the testing subsets is presented in Table \ref{tab:static_test_set}.

\begin{table*}
    \centering
    \fontsize{8}{10}\selectfont
    \caption{
    Performance comparison of baseline and state-of-the-art person ReID methods on different datasets: Market-1501 (Ground-Ground), UAV-Human (Aerial-Aerial), and AG-ReID.v2 (Aerial-Ground). In this context, \textbf{G} represents \textbf{Ground}, \textbf{A} for \textbf{Aerial}, \textbf{C} for \textbf{CCTV}, and \textbf{W} for \textbf{Wearable}. {The AG-ReID.v2 dataset tests include cross-domain adaptations:} \textbf{A $\rightarrow$ C} (Aerial to CCTV), \textbf{A $\rightarrow$ W} (Aerial to Wearable), \textbf{C $\rightarrow$ A} (CCTV to Aerial), and \textbf{W $\rightarrow$ A} (Wearable to Aerial).}
    
    \begin{tabular}{lcccccccccccc}
    \toprule
        \multirow{3}{*}{\textbf{Model}}
        & \multicolumn{2}{c}{\textbf{G $\rightarrow$ G}}
        & \multicolumn{2}{c}{\textbf{A $\rightarrow$ A}}
        & \multicolumn{2}{c}{\textbf{A $\rightarrow$ C}}
        & \multicolumn{2}{c}{\textbf{A $\rightarrow$ W}}
        & \multicolumn{2}{c}{\textbf{C $\rightarrow$ A}}
        & \multicolumn{2}{c}{\textbf{W $\rightarrow$ A}}
        \\ \cmidrule(lr){2-3} \cmidrule(lr){4-5} \cmidrule(lr){6-7} \cmidrule(lr){8-9} \cmidrule(lr){10-11} \cmidrule(lr){12-13}
        
        & mAP & Rank-1 & mAP & Rank-1 & mAP & Rank-1 & mAP & Rank-1 & mAP & Rank-1 & mAP & Rank-1
        \\ \midrule
        
        Swin \cite{Liu2021SwinTH}
        & 79.70 & 92.75 & 67.37 & 68.23 & 57.66 & 68.76 & 56.15 & 68.49 & 57.70 & 68.80 & 53.90 & 64.40
        \\ 

        HRNet-18 \cite{Wang2019DeepHR}
        & 76.65 & 90.83 & 64.52 & 65.48 & 65.07 & 75.21 & 66.17 & 76.26 & 66.16 & 76.25 & 66.17 & 76.25
        \\ 

        SwinV2 \cite{liu2021swinv2}
        & 82.99 & 92.93 & 69.15 & 70.12 & 66.09 & 76.44 & 69.09 & 80.08 & 62.14 & 77.11 & 65.61 & 74.53
        \\

        MGN (R50) \cite{Wang2018LearningDF}
        & 86.90 & 95.70 & 70.40 & 70.38 & 70.17 & {82.09} & 78.66 & 88.14 & 72.41 & {84.21} & 73.73 & 84.06
        \\ 

        BoT (R50) \cite{Luo2019BagOT}
        & 83.95 & 94.77 & 63.41 & 62.48 & 71.49 & 80.73 & 75.98 & 86.06 & 69.67 & 79.46 & 72.41 & 82.69
        \\ 
        
        BoT (R50) + Attributes
        & 84.92 & 95.37 & 64.11 & 63.28 & 72.19 & 81.43 & 76.68 & 86.66 & 70.37 & 80.15 & 73.11 & 83.29
        \\

        SBS (R50) \cite{He2020FastReIDAP}
        & 88.20 & 95.40 & 65.93 & 66.38 & 72.04 & 81.96 & 78.94 & 88.14 & 73.89 & 84.10 & 75.01 & 84.66
        \\
        
        SBS (R50) + Attributes
        & 88.90 & 96.00 & 66.63 & 67.38 & 72.74 & 82.56 & 79.64 & 88.74 & 74.59 & 84.80 & 75.71 & 85.26
        \\ \hline

        \textbf{V2E (ViT) - Ours}& \textbf{90.25} & \textbf{96.35} & \textbf{71.47} & \textbf{72.75} & \textbf{80.72} & \textbf{88.77} & \textbf{84.85} & \textbf{93.62} & \textbf{78.51} & \textbf{87.86} & \textbf{80.11} & \textbf{88.61}
        \\
        \bottomrule
    \end{tabular}
    \label{comp_sota_dataset}
\end{table*}

\begin{figure}
    \centering
    \begin{multicols}{2}
      
        \includegraphics[width=0.9\columnwidth, height=6cm]{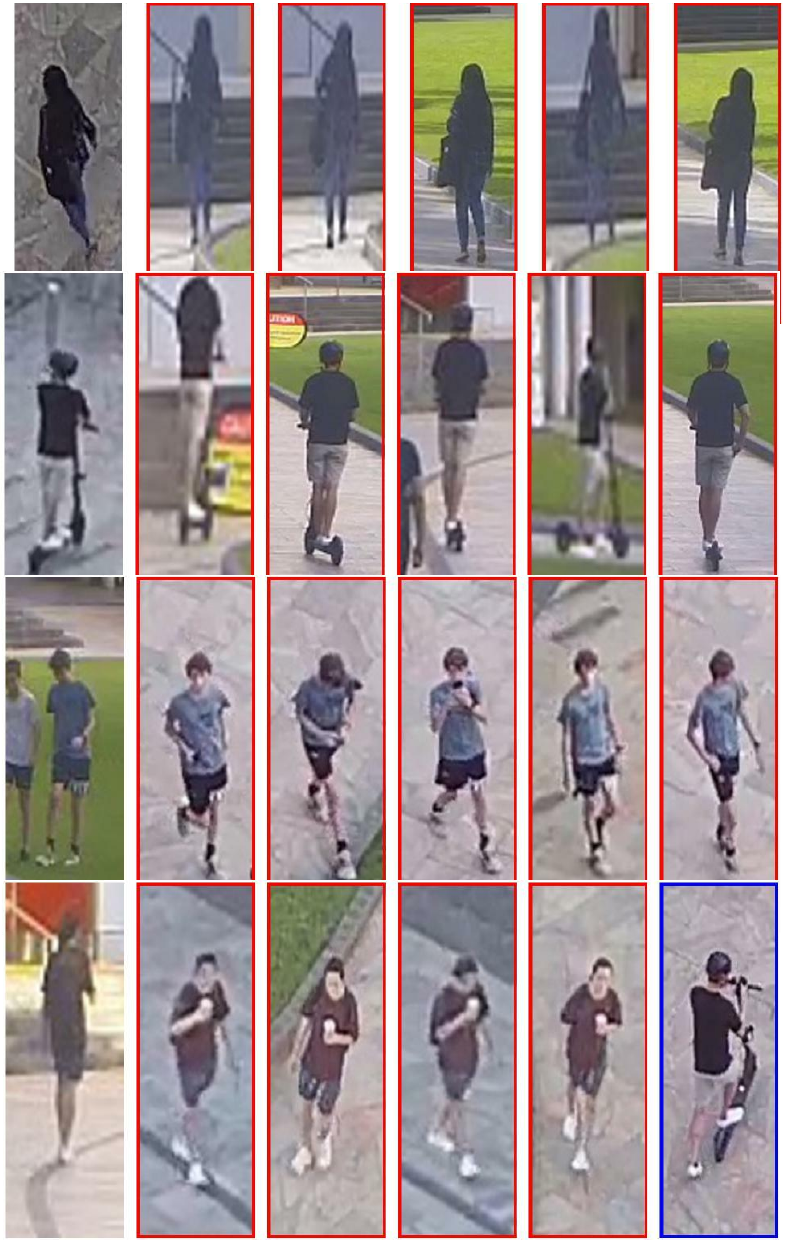} \\
        \small (a) Successful ReID Matches
      
        \includegraphics[width=0.9\columnwidth, height=6cm]{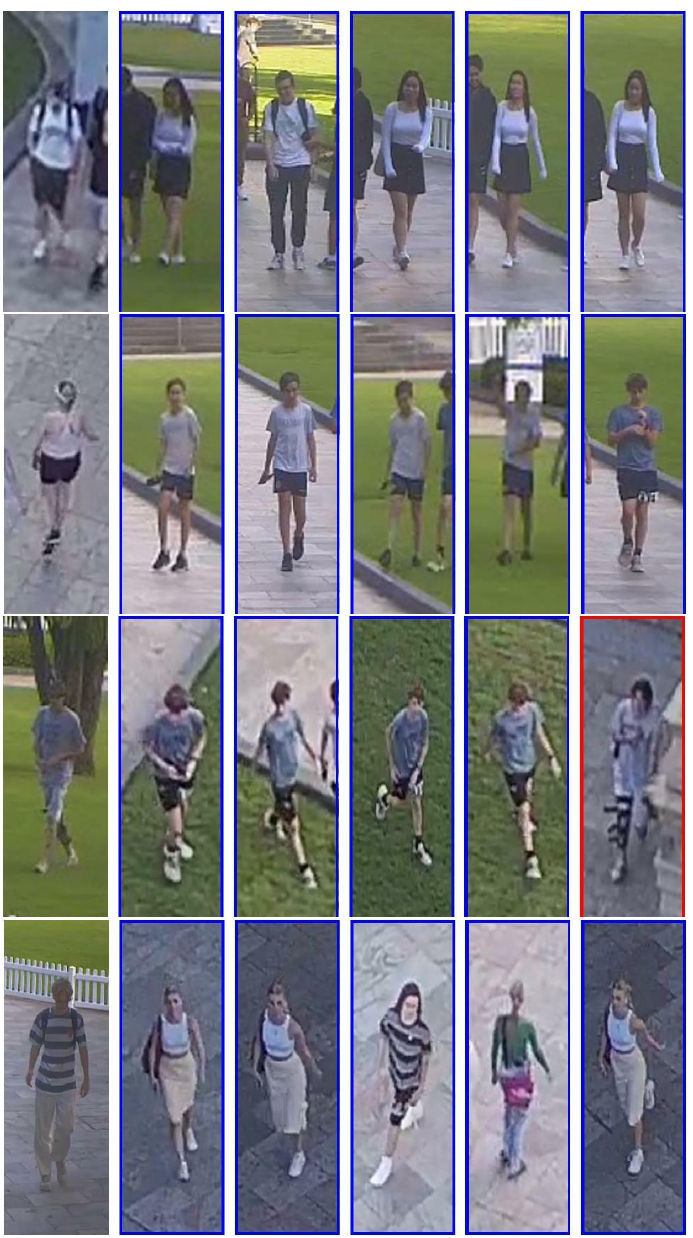} \\
        \small (b) ReID Mismatches
             \label{fig:ranking_failure}
      
    \end{multicols}
    
    \caption{Examples highlighting our model's performance on AG-ReID.v2.}
    \label{fig:combined_ranking}
    
\end{figure}

\subsection{{Implementation and Evaluation Metrics}}\label{sec:Implementation and Evaluation Metrics}

In our study, we employed established person re-identification (ReID) techniques, including BoT Baseline \cite{Luo2019BagOT}, StrongerBaseline (SBS) \cite{He2020FastReIDAP}, MGN \cite{Wang2018LearningDF}, HRNet-18 \cite{Wang2019DeepHR}, Swin \cite{Liu2021SwinTH}, and SwinV2 \cite{liu2021swinv2}. These ReID models were built on various backbones such as ResNet-50, Vision Transformer (ViT), OSNet, HRNet-18, Swin, and SwinV2, each pre-trained on the ImageNet dataset \cite{Deng2009ImageNetAL}. The models using ResNet-50 and OSNet backbones were optimized with the Adam algorithm \cite{Kingma2015AdamAM} at a learning rate of $10^{-4}$. In contrast, models with HRNet-18, Swin, and SwinV2 backbones utilized the SGD optimizer at a rate of $5^{-2}$. The ViT backbone model was trained with the SGD optimizer at a learning rate of $10^{-3}$.

\noindent{%
\begin{minipage}{\columnwidth}
\hspace{0.25cm} Additionally, the component $\mathcal{G}(\cdot)$, functioning as a crucial part of our V2E network, was rigorously trained on the Market-1501, UAV-Human, and AG-ReID.v2 datasets. For these datasets, the number of attributes, denoted as $M$, were established at $28$, $38$, and $88$, respectively. The hyper-parameters, specified in Equation \ref{eq:total_loss}, were set as $\alpha = 10.0$ and $\beta = 50.0$. Additionally, the parameter $\upsilon$ in Equation \ref{equ:equ14} was determined to be $0.5$, while  \(\mathcal{K}\)  and $\mathcal{T}$ in Equation \ref{equ:equ6} were assigned values of $1/M$ and $0.5$, respectively. The chosen mini-batch size for the training was configured as $6 \times 4$, implying the inclusion of $6$ IDs, each with $4$ samples. In line with the StrongerBaseline (SBS) method \cite{He2020FastReIDAP}, our V2E model integrates shared convolutional layers with the main ReID model. This integration facilitates effective pattern recognition, particularly for elevated camera angles, and enables comprehensive performance analysis under varying camera perspectives. By sharing convolutional layers between the ReID and V2E models, we achieved a reduction in computational demands during inference, thereby enhancing the overall efficiency and interpretability of our ReID model.
\end{minipage}%
}

For evaluation, we employed the mean Average Precision (mAP) \cite{Zheng2015ScalablePR} and Cumulative Matching Characteristics (CMC-$k$) \cite{Wang2007ShapeAA} metrics. The mAP metric averages the precision across different recall levels, considering multiple ground truth labels, while the CMC-$k$ metric gauges the likelihood of correct identification within the top-k results, with our report focusing on rank-1 performance.

\subsection{Comparison with State-of-the-art Models} \label{sec:Comparison with State-of-the-art Models}

In this section, a comparative analysis is conducted to assess the efficacy of various Re-identification (ReID) models on three datasets, summarized in Table \ref{comp_sota_dataset}. We begin with the Strong Baseline (BoT) model which is built upon the ResNet50 architecture, as described in \cite{Luo2019BagOT}.

In the Aerial to CCTV dataset, the BoT model achieved an mAP of 71.49\% and a Rank-1 accuracy of 80.73\%. When evaluated on the Aerial to Wearable dataset, there was an improvement with an mAP of 75.98\% and Rank-1 accuracy of 86.06\%. In the Ground to Aerial scenario, which considers queries from both CCTV and wearable platforms, the BoT model reported an mAP of 69.67\% and a Rank-1 accuracy of 79.46\% for the CCTV-Aerial pairing, and an mAP of 72.41\% with a Rank-1 accuracy of 82.69\% for the Wearable-Aerial configuration. Notably, on the Market-1501 dataset, which is solely ground-based, the BoT achieved an mAP of 83.95\% and a Rank-1 accuracy of 94.77\%. This disparity accentuates the inherent complexities of ReID involving aerial data.

Further analyses were conducted on models such as HRNet-18 and SwinV2. HRNet-18 \cite{Wang2019DeepHR} produced satisfactory results but remained behind BoT in the Aerial to CCTV and Aerial to Wearable datasets, with mAP values of 65.07\% and 66.17\%, and Rank-1 accuracies of 75.21\% and 76.26\%, respectively. SwinV2 \cite{liu2021swinv2} was more competitive, achieving an mAP of 66.09\% and a Rank-1 accuracy of 76.44\% in the Aerial to CCTV dataset, and an mAP of 69.09\% with a Rank-1 accuracy of 80.08\% in the Aerial to Wearable dataset. A detailed comparison with Swin \cite{Liu2021SwinTH} can offer further insights into the differential performance metrics of SwinV2.

Our proposed V2E model presented noteworthy advancements. In the Aerial to CCTV dataset, V2E recorded an mAP of 80.72\% and a Rank-1 accuracy of 88.77\%. This trend persisted in the Aerial to Wearable dataset where V2E achieved an mAP of 84.85\% and a Rank-1 accuracy of 93.62\%.

\subsection{Ablation Studies}\label{sec:Explainable Elevated-view Attention method on AG-ReID.v2 Dataset}

To offer a more comprehensive understanding of the contribution of each component, Table \ref{tab:interpreter_reweight_combined} presents the outcomes of our Explainable Elevated-View Attention (EP+EVA) technique on the AG-ReID.v2 dataset.

The synergy of the Vision Transformer (ViT) with the Explainable (EP) component manifested a marked improvement in the aerial-to-CCTV setting. The mAP surged from 77.03\% to 79\%, registering a growth of 1.97\%. Concurrently, the Rank-1 accuracy experienced a 2.3\% increment, escalating from 85.4\% to 87.7\%.

Expanding upon the role of each component:
\begin{enumerate}
    \item \textbf{Vision Transformer (ViT) Backbone}: The ViT backbone is a key component of our architecture. Our ablation study demonstrates its effectiveness, particularly in its compatibility with EP and EVA components. This integration contributes to enhanced re-identification performance in diverse scenarios.
    \item \textbf{Elevated-View Attention (EVA) Mechanism}: Incorporating the EVA mechanism with the ViT backbone resulted in a measurable improvement in performance metrics: an increase in mAP by 1.67\%, reaching 78.70\%, and a rise in Rank-1 accuracy by 1.33\%, achieving 86.73\%. The EVA mechanism effectively addresses challenges specific to aerial imagery and refines reID methods through optimized partition and attention layers.
    \item \textbf{Explainable Stream’s Attention Map}: The attention map within the Explainable Stream is designed to locate distinct attributes or features. This targeted focus enhances the model's capability to provide insights and improves overall performance, distinguishing it from conventional spatial attention mechanisms.
\end{enumerate}

\begin{table*}
    \centering
    \fontsize{8}{10}\selectfont
     \caption{Performance of the proposed explainable elevated-view attention method on AG-ReID.v2 dataset. \textbf{ViT} denotes Vision Transformer backbone, \textbf{EP} denotes explainable processing, \textbf{EVA} denotes elevated-view attention. {Performance improvements over ViT are indicated with \textbf{bold text}, with the highest improvement highlighted in \textcolor{red}{\textbf{red}}.}}
    \label{tab:interpreter_reweight_combined}

    \begin{tabular}{lcccccccc}
        \toprule
        \multirow{3}{*}{\textbf{Model}}
        & \multicolumn{2}{c}{\textbf{Aerial $\rightarrow$ CCTV}}
        & \multicolumn{2}{c}{\textbf{Aerial $\rightarrow$ Wearable}}
        & \multicolumn{2}{c}{\textbf{CCTV $\rightarrow$ Aerial}}
        & \multicolumn{2}{c}{\textbf{Wearable $\rightarrow$ Aerial}} \\
        \cmidrule(lr){2-3} \cmidrule(lr){4-5} \cmidrule(lr){6-7} \cmidrule(lr){8-9}
        & mAP & {Rank-1} 
        & mAP & {Rank-1} 
        & mAP & {Rank-1} 
        & mAP & {Rank-1} \\
        \midrule
        {ViT}
        & 77.03 & 85.40
        & 80.48 & 89.77
        & 75.90 & 84.65
        & 76.59 & 84.27 \\ 
        {ViT+EP}
        & 79(\textbf{+1.97}) & 87.7(\textbf{+2.3})
        & 83.14(\textbf{+2.66}) & 93.67(\textbf{+3.9})
        & 78.24(\textbf{+2.34}) & 87.35(\textbf{+2.7})
        & 79.08(\textbf{+2.49}) & 87.73(\textbf{+3.46}) \\
        {ViT+EVA}
        & 78.70(\textbf{+1.67}) & 86.73(\textbf{+1.33})
        & 81.97(\textbf{+1.49}) & 90.5(\textbf{+0.73})
        & 76.23(\textbf{+0.33}) & 85.1(\textbf{+0.45})
        & 77.8(\textbf{+1.21}) & 85.83(\textbf{+1.56}) \\
        {ViT+EVA+EP}
        & 80.72(\textbf{+3.69}) & 88.77(\textbf{+3.37})
        & 84.85({\textcolor{red}{\textbf{+4.37}}}) & 93.62(\textbf{+3.85})
        & 78.51(\textbf{+2.61}) & 87.86(\textbf{+3.21})
        & 80.11(\textbf{+3.52}) & 88.61(\textbf{+4.34}) \\
        \bottomrule
    \end{tabular}
\end{table*}

\begin{figure}
    \centering
	\includegraphics[width=0.9\columnwidth]{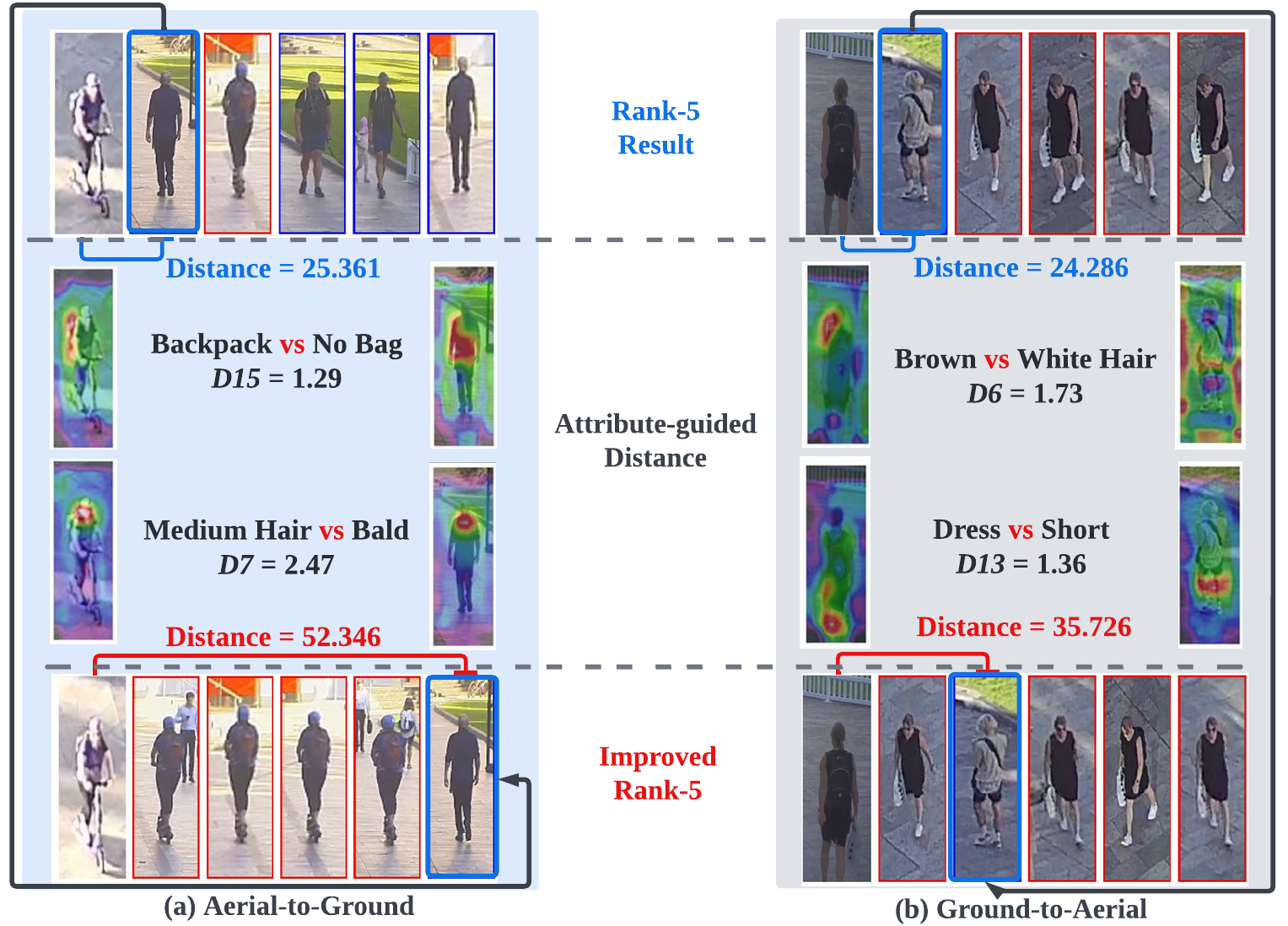}
	\caption{Model's improved Rank-1 accuracy: aerial-to-ground (left) vs. ground-to-aerial (right), influenced by attributes.}
    \label{fig:vis_explainable_rank1}
\end{figure}

Our results are visually represented in Figures \ref{fig:combined_ranking} and \ref{fig:vis_explainable_rank1}, which complement our analytical findings. Figure \ref{fig:combined_ranking} displays query images alongside their top-5 gallery matches, with correct matches bordered in red and mismatches in blue. This figure highlights both accurate and erroneous matches in the aerial-to-CCTV context (top two rows), and in the CCTV-to-aerial context (bottom rows). Common causes of misidentification, including similar clothing, postures, and camera perspectives, are evident in these examples. In contrast, Figure \ref{fig:vis_explainable_rank1} illustrates the impact of different attributes on the accuracy of our model in both aerial-to-ground and ground-to-aerial scenarios.

\subsection{Discussion}

Our study extends the foundational work of Chen et al. \cite{Chen2021ExplainablePR} by introducing a transformer-based architecture tailored for aerial-ground person re-identification. As detailed in Stream 1 \ref{subsec:Stream 1: Transformer-based Person ReID}, our approach surpasses the limitations of Chen et al.’s methodology \cite{Chen2021ExplainablePR} in processing low-resolution images, a common challenge in aerial imagery. This capability significantly enhances the model's effectiveness in aerial-ground ReID scenarios.

In terms of performance, our model, featuring the V2E network, demonstrates substantial improvements in both mAP and Rank-1 accuracy, particularly in aerial to CCTV and aerial to wearable datasets. This indicates a marked advancement in the model's ability to handle complex aerial-ground ReID tasks.

Distinct from Chen et al. \cite{Chen2021ExplainablePR}, our model incorporates several innovations: the use of 2D Adaptive Average Pooling for improved feature extraction, Automatic Mixed Precision (AMP) Training with the GradScale method for optimized computational efficiency and precision, and the training on 88 Binary Vector Attributes derived from 15 soft attribute labels for a more detailed identity analysis. These features collectively enhance the model's robustness and accuracy, setting it apart from previous methodologies.

Finally, our comprehensive evaluations of various backbones, including ResNet, ViT, and OSNet, led to the selection of ViT as the most suitable for our dataset and application. This choice highlights the model's adaptability and effectiveness in aerial-ground ReID applications.

\section{Conclusion}\label{Conclusion}

In this paper, we present a novel contribution to Aerial-Ground ReID research through the introduction of an expanded dataset, building upon our prior work. This dataset is unique in its integration of diverse data sources: aerial, CCTV, and, notably, wearable technologies. It comprises 1,615 distinct person identities represented by a total of 100,502 images. To effectively tackle the specific challenges introduced by wearable data, our approach incorporates a three-stream attribute-based ReID methodology. This methodology is distinguished by an explainable elevated-view attention mechanism, primarily focused on the head region and attribute attention maps. This strategic emphasis enables our approach to overcome traditional limitations in ReID tasks. Our methodology demonstrates marked improvements in rank accuracy and overall performance. In line with our commitment to supporting ongoing research in this field, we will make the enriched dataset and the baseline code freely available to the research community.

\section*{Acknowledgments}

This study, funded by the Australian Research Council (ARC) Discovery (Project No. DP200101942) and a QUT Postgraduate Research Award, acknowledges the support of the Research Engineering Facility (REF) team at QUT for their expertise and research infrastructure, which have been instrumental in enabling this project. Ethics approval for this study has been secured, and to ensure privacy, participant facial regions were pixelated.


\bibliographystyle{IEEEtran}
\bibliography{tifs}

%

\begin{IEEEbiography} [{\includegraphics[width=1in,height=1.25in,clip,keepaspectratio]{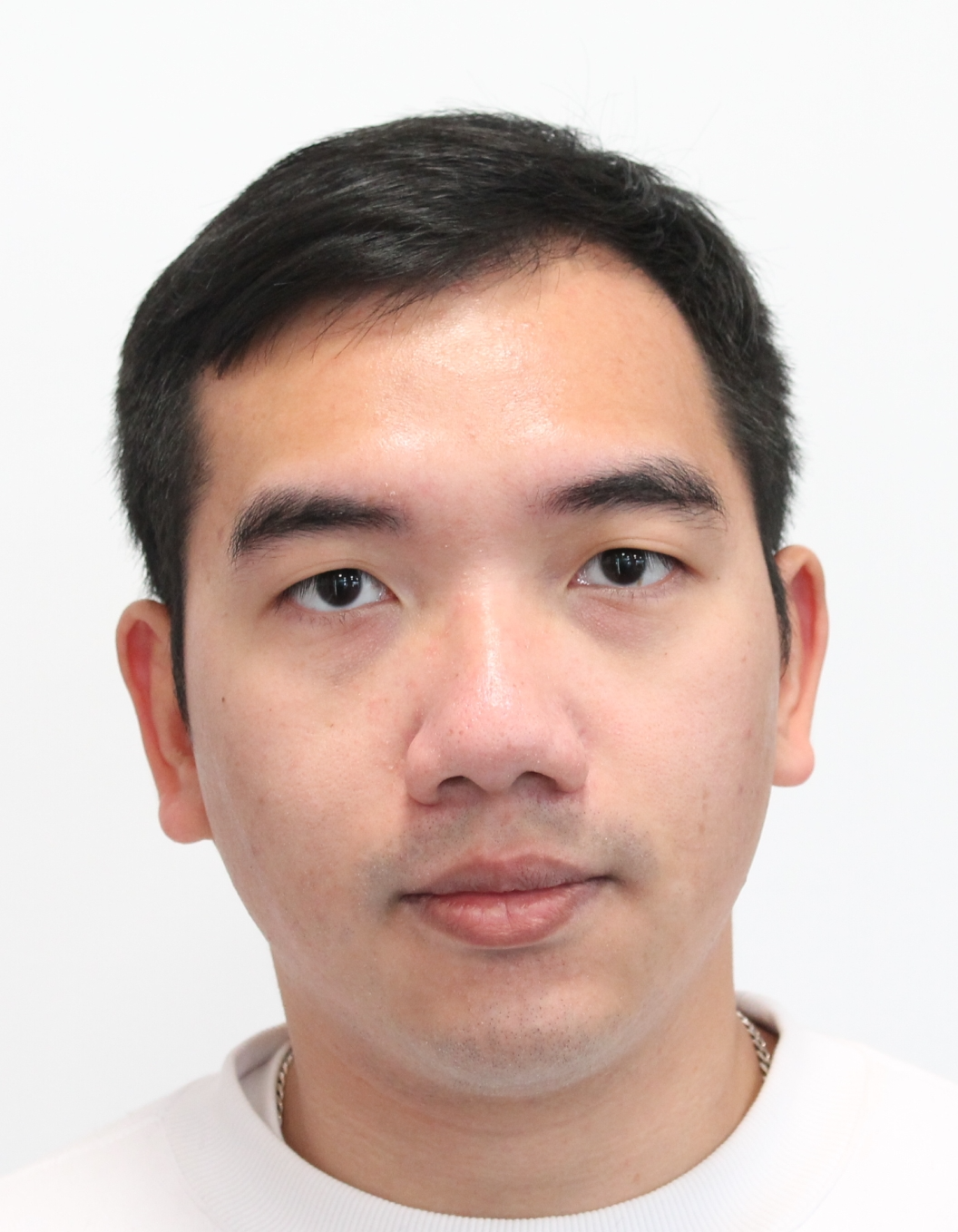}}]{Huy Nguyen}
received the B.Eng. (Hons) degree in Computer and Software Systems from Queensland University of Technology (QUT). Currently, he is pursuing a Ph.D. in computer vision at QUT. His research interests include computer vision, aerial surveillance and its application using deep learning models.
\end{IEEEbiography}

\begin{IEEEbiography}
    [{\includegraphics[width=1in,height=1.25in,clip,keepaspectratio]{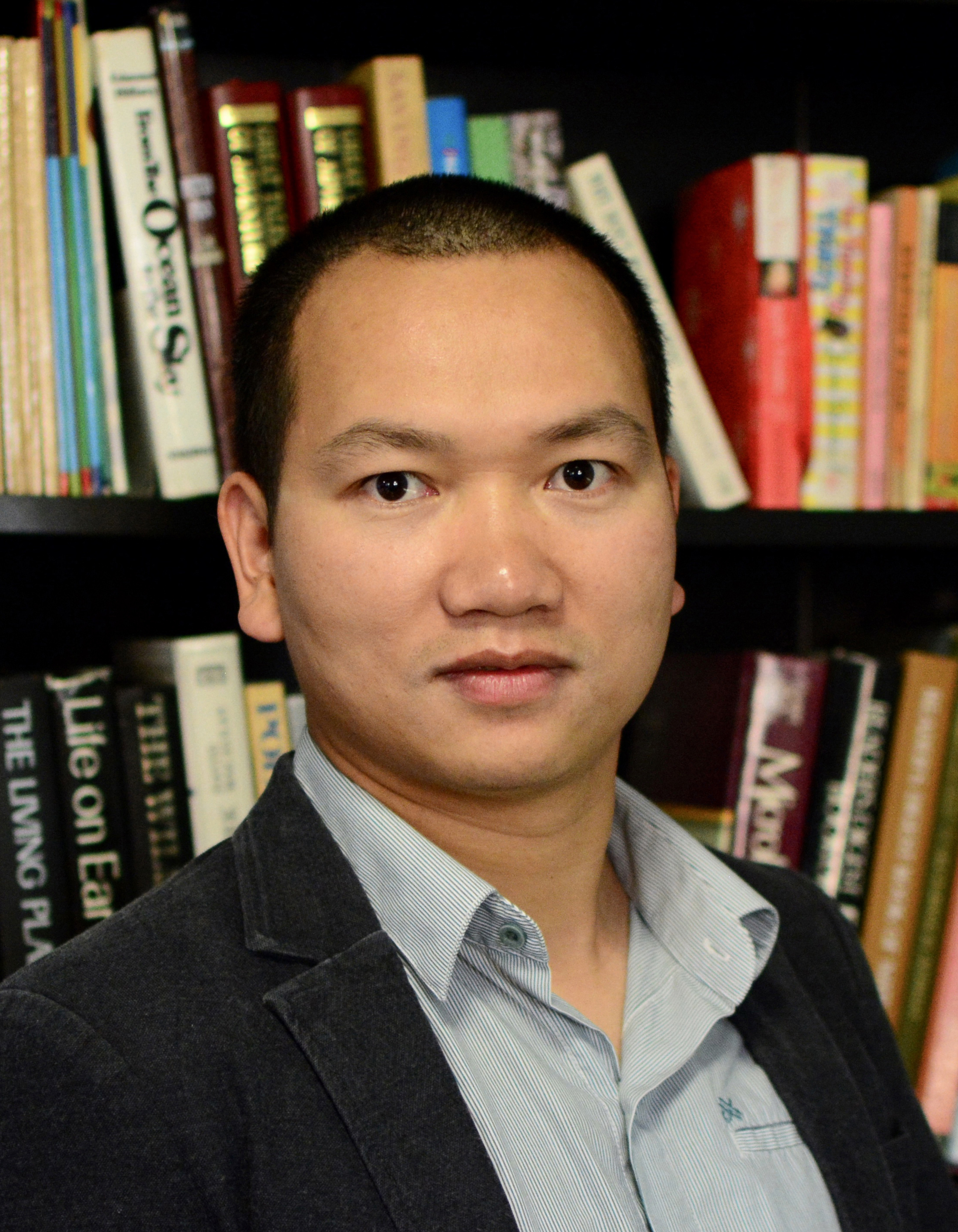}}]{Kien Nguyen}
is a Senior Research Fellow at Queensland University of Technology. He has been conducting research in surveillance and biometrics for the last 10 years, and has published his research in high quality journals and conferences in the area. His research interests are in application of computer vision and deep learning techniques to the areas of biometrics, surveillance and scene understanding. He has been serving as an Associate Editor for IEEE Access, and Image and Vision Computing.
\end{IEEEbiography}

\begin{IEEEbiography}
    [{\includegraphics[width=1in,height=1.25in,clip,keepaspectratio]{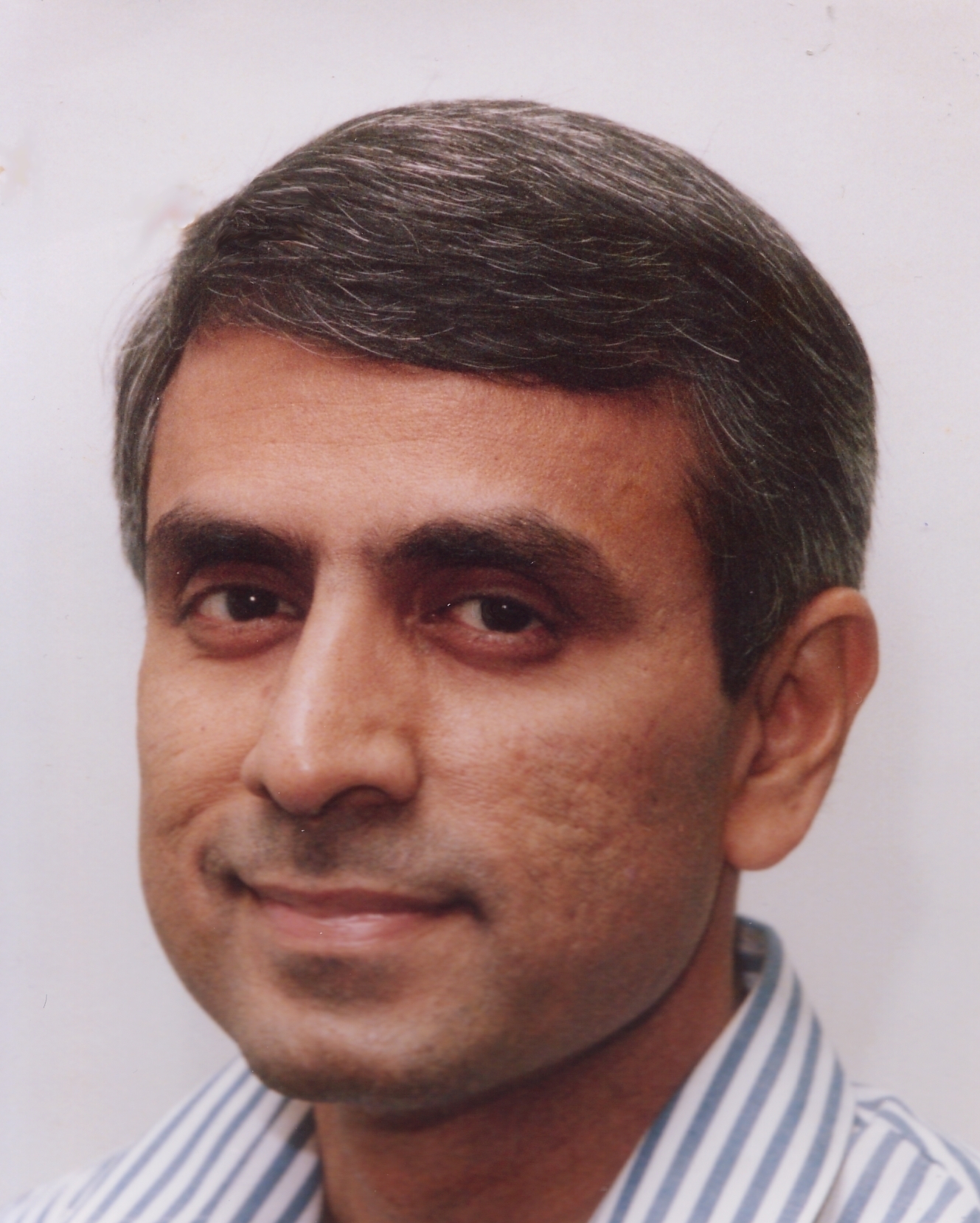}}]{Sridha Sridharan} obtained his MSc degree from the University of Manchester, UK and his PhD degree from University of New South Wales, Australia. He is currently a Professor at Queensland University of Technology (QUT) where he leads the research program in Signal Processing, Artificial Intelligence and Vision Technologies (SAIVT).
\end{IEEEbiography}

\begin{IEEEbiography}
    [{\includegraphics[width=1in,height=1.25in,clip,keepaspectratio]{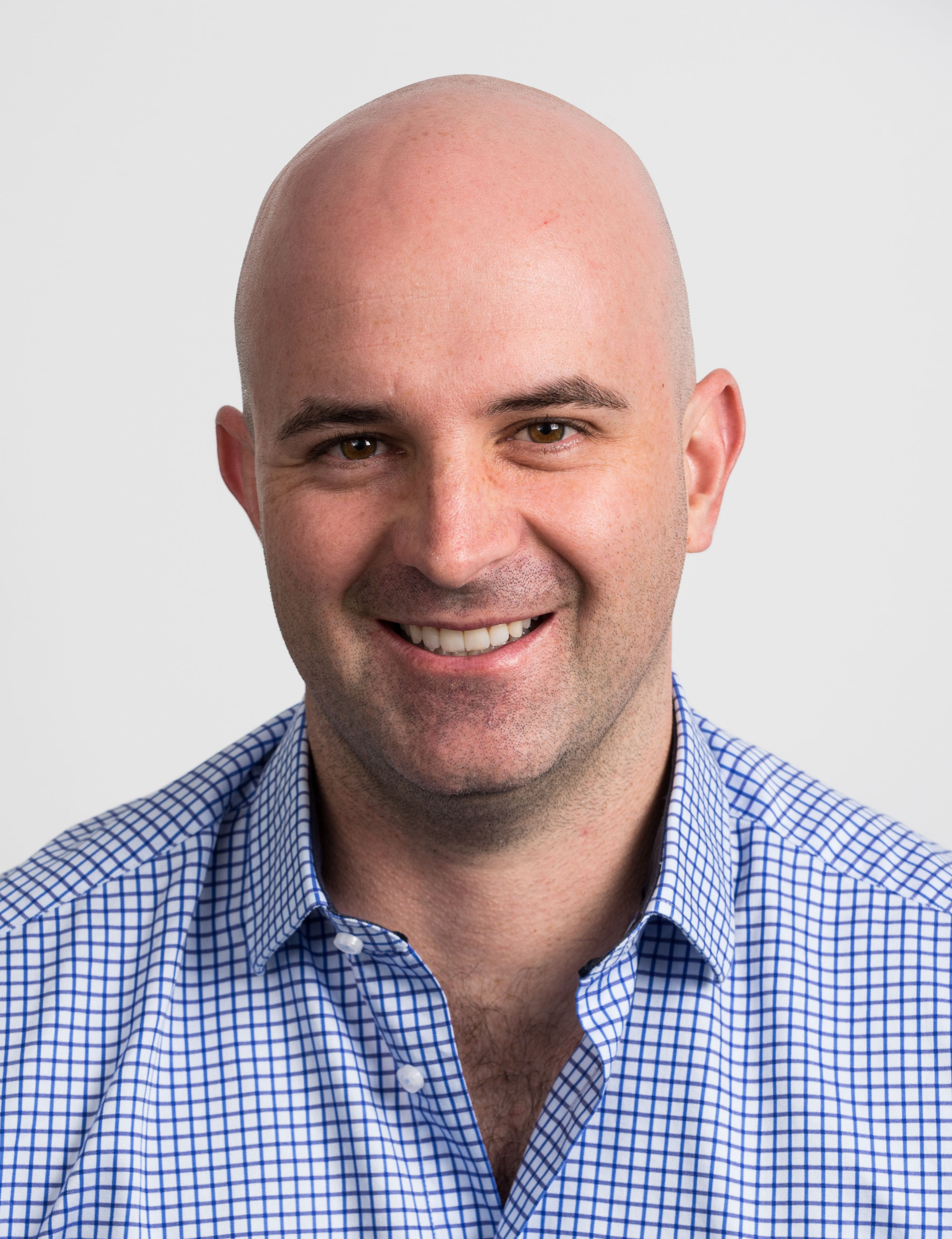}}]{Clinton Fookes} is a Professor in Vision and Signal Processing at the Queensland University of Technology. He holds a BEng (Aero/Av), an MBA, and a PhD in computer vision. He actively researches across computer vision, machine learning, signal processing and pattern recognition areas. He serves on the editorial boards for the IEEE Transactions on Image Processing, Pattern Recognition, and the IEEE Transactions on Information Forensics and Security. He is a Senior Member of the IEEE, an Australian Institute of Policy and Science Young Tall Poppy, an Australian Museum Eureka Prize winner, and a Senior Fulbright Scholar.

\end{IEEEbiography}







\end{document}